%% file: main.tex
\documentclass[sigconf]{acmart}

\AtBeginDocument{%
  \providecommand\BibTeX{{%
    \normalfont B\kern-0.5em{\scshape i\kern-0.25em b}\kern-0.8em\TeX}}}

\copyrightyear{2020}
\acmYear{2020}
\setcopyright{rightsretained}
\acmConference[KDD '20]{Proceedings of the 26th ACM SIGKDD Conference on Knowledge Discovery and Data Mining}{August 23--27, 2020}{Virtual Event, CA, USA}
\acmBooktitle{Proceedings of the 26th ACM SIGKDD Conference on Knowledge Discovery and Data Mining (KDD '20), August 23--27, 2020, Virtual Event, CA, USA}
\acmDOI{10.1145/3394486.3403153}
\acmISBN{978-1-4503-7998-4/20/08}

\usepackage[normalem]{ulem}
\usepackage{amsmath}
\usepackage[linesnumbered,algoruled,boxed,noend]{algorithm2e}
\useunder{\uline}{\ul}{}
\begin{document}

\newtoggle{showtodos}
\toggletrue{showtodos}
\newcommand{\todo}[3]{\iftoggle{showtodos}{\textcolor{#1}{\textbf{#2: #3}}}{}}

\newcommand{\tata}[1]{\todo{blue}{Sandeep}{#1}}
\newcommand{\yuchen}[1]{\todo{blue}{Yuchen}{#1}}
\newcommand{\highlight}[1]{{\color{purple}{#1}}}
\newcommand{\ying}[1]{\todo{red}{Ying}{#1}}
\newcommand{\nguyenvo}[1]{\todo{red}{Nguyen}{#1}}
\newcommand{\eat}[1]{}


\title{FreeDOM: A Transferable Neural Architecture for \\ Structured Information Extraction on Web Documents}


\author{Bill Yuchen Lin}
\authornote{The work was done while BYL was a research intern at Google AI.}
\affiliation{%
  \institution{University of Southern California}
  \city{Los Angeles}
  \state{California}
  \country{USA}
}
\email{yuchen.lin@usc.edu}

\author{Ying Sheng}
\affiliation{%
  \institution{Google}
  \city{Mountain View}
  \state{California} 
  \country{USA}
}
\email{yingsheng@google.com}

\author{Nguyen Vo}
\affiliation{%
  \institution{Google}
  \city{Mountain View}
  \state{California} 
  \country{USA}
}
\email{nguyenvo@google.com}

\author{Sandeep Tata}
\affiliation{%
  \institution{Google}
  \city{Mountain View}
  \state{California} 
  \country{USA}
}
\email{tata@google.com}

\fancyhead{}

\begin{abstract}

Extracting structured data from HTML documents is a long-studied problem with a broad range of applications like augmenting knowledge bases, supporting faceted search, and providing domain-specific experiences for key verticals like shopping and movies.
Previous approaches have either required a small number of examples for each target site or relied on carefully handcrafted heuristics built over visual renderings of websites.
In this paper, we present a novel two-stage neural approach, named \textsc{FreeDOM}, which overcomes both these limitations. 
The first stage learns a representation for each DOM node in the page by combining both the text and markup information.
The second stage captures longer range distance 
and semantic relatedness using a relational neural network.
By combining these stages, \textsc{FreeDOM} is able to generalize to unseen sites after training on a small number of seed sites from that vertical without requiring expensive hand-crafted features over visual renderings of the page.
Through experiments on a public dataset with 8 different verticals, we show that \textsc{FreeDOM} beats the previous state of the art by nearly 3.7 F1 points on average \emph{without} requiring features over rendered pages or expensive hand-crafted features.


\end{abstract}




\maketitle
\input{sec_1_intro}
\input{sec_2_problem}

\input{sec_3_overview}
\input{sec_4_node}

\input{sec_5_relation}

\input{sec_6_exp}
\input{sec_7_related}

\section{Conclusions and Future Work}
\label{sec:conclusion}
In this paper, we propose a neural architecture for extracting structured data from web documents.
It uses training data from only a few seed sites but generalizes well to other unseen websites in the same vertical.
We show that our approach,
\textsc{FreeDOM}, beats the previous state-of-the-art performance on a large-scale public dataset consisting of 8 different verticals (80 websites in total) by nearly 3.7 F1 points. In particular, it does so without using any expensive rendering-based visual features.

We believe that this work opens up multiple avenues for future research in web data extraction.
What structured prediction techniques might work better at incorporating information from farther away and work well on large DOM trees with sparse labels? 
An even more interesting question is if we can transfer information across verticals? That is, if we are able to do well on one vertical, can we leverage that information somehow to train a model for the next vertical~\cite{Lockard2020ZeroShotCeresZR}?
Will there be a large pre-trained neural encoding model for HTML documents, just like BERT~\cite{Devlin2019BERTPO} for plain texts?

\bibliographystyle{plainnat}
\bibliography{sample-base}
\clearpage
\input{sec_8_supp}

\end{document}

%% file: sec_1_intro.tex
\section{Introduction} 
\begin{figure}[t]
	\centering
	\includegraphics[width=1\linewidth]{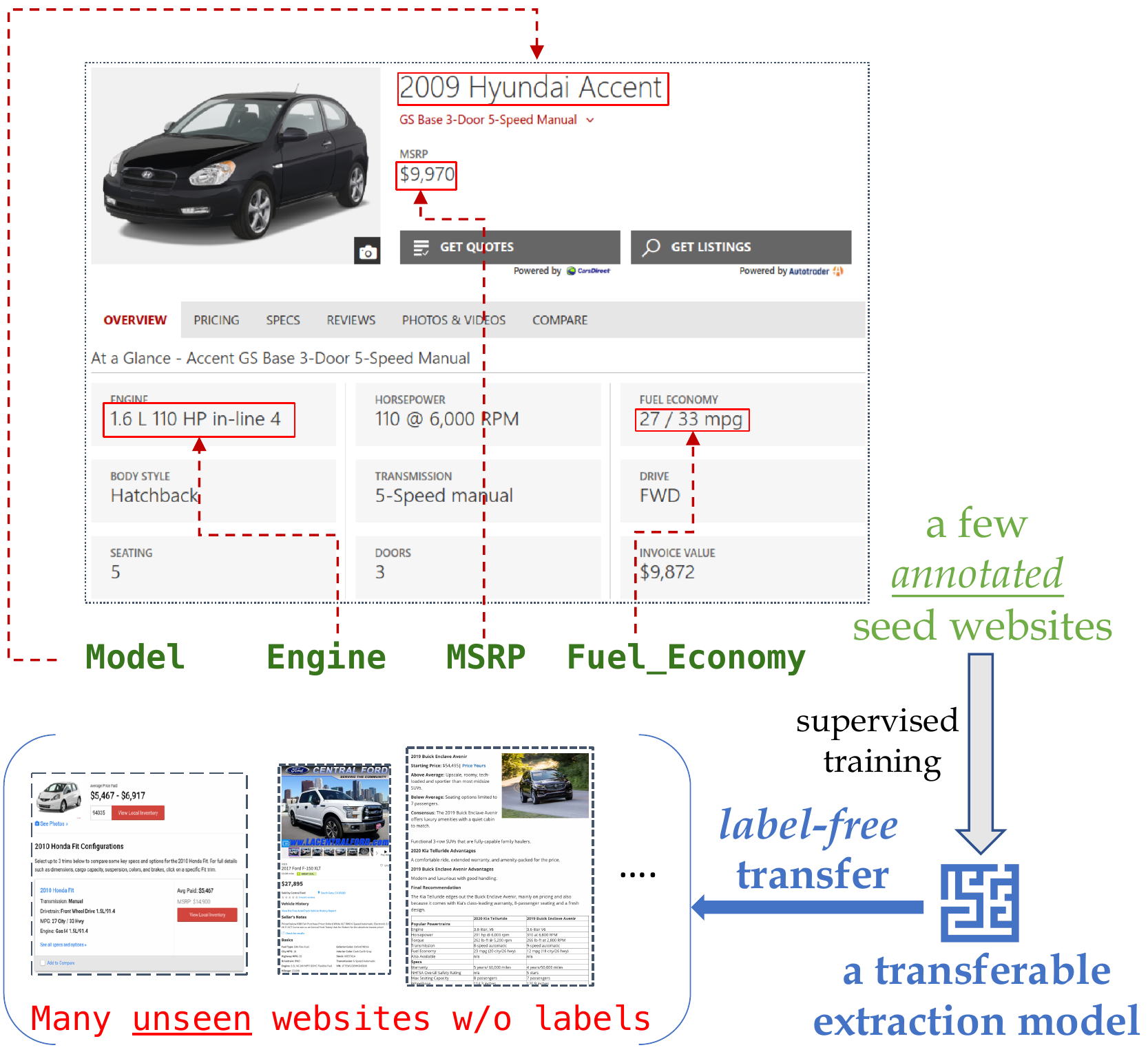}
	\caption{Learning a transferable extraction model that supports label-free adaption on unseen websites.\vspace{-15pt}
	}
	\Description{}
	\label{fig:intro}
	
\end{figure}

Extracting structured information from web-pages is critical for large-scale knowledge base construction~\cite{Chang2006ASO,Dong2014KnowledgeVA, Dong2014FromDF,Wu2018FonduerKB}.
This in turn has various downstream applications such as knowledge-aware question answering~\cite{Cui2017KBQALQ, talmor-berant-2018-web,lin-etal-2019-kagnet}, recommendation systems~\cite{Wang2019MultiTaskFL, Ma2019JointlyLE,Luo2018ExtRAEP}, temporal event reasoning~\cite{Trivedi2017KnowEvolveDT, mlg2019_45}, etc.
Most recent advances in information extraction focus on neural models for extracting entities and relations on plain texts (i.e. natural language sentences)~\cite{Ren2016AFETAF, NERO2020}.
However, these neural information extraction methods do not work well on web documents which are richly laid out and contain both natural language and markup information.
%

In this paper, we focus on extracting structured data with given attributes from \emph{detail pages}.
A detail page describes a single entity such as an IMDB page of a particular movie or an Amazon page about a certain product.
For instance, consider that we want to extract information about vehicles from various websites and are interested in four attributes: \{\textit{model}, \textit{MSRP}, \textit{engine}, \textit{fuel-economy}\}.
Conventional web data extraction methods are based on ``wrapper induction''~\cite{Kushmerick1997WrapperIF}.
For each website, they require a handful of human-annotated pages and derive a pattern-based program that can work on other pages within the \emph{same} website.
This approach yields very high accuracy (over 95\%~\cite{Gulhane2011WebscaleIE}). 
However, these wrapper-based methods need human annotations for each new website. This makes them expensive, time-consuming, and thus less practical when we have a large number of target websites of interest.
In this paper, we are interested in an approach where we can learn from only a few annotated seed websites but then generalize to many unseen sites without additional labels (Figure~\ref{fig:intro}).

%

Prior work primarily focused on exploiting visual patterns using carefully crafted features~\cite{Hao2011FromOT} .
These rendering-based methods have two major drawbacks: 1) they are expensive since they require downloading all external files including CSS, javascript, and images to render the page to compute visual features; 2) they require carefully crafted heuristics around visual proximity  to work well with these expensive features.
In this paper, we propose a novel two-stage neural architecture, named \textsc{FreeDOM}, that can be trained on a small number of seed websites and generalize well to unseen websites \emph{without} requiring any hand-engineered visual features.

The \textit{first stage} of our framework (Sec.~\ref{ssec:node_module}) learns a representation for each node in the DOM tree\footnote{{A DOM Tree, associated with a web page, is a collection of nodes, where each node has its address named XPath and original HTML content.}} of the page and classifies it into one of the target fields.
This node-level module combines neighboring character sequences, token sequences, as well as markup (HTML elements) to learn a combined representation for the node.
We propose a combination of CNNs and LSTMs and show that it can effectively encode useful features in DOM nodes.

These node representations are encoded individually and inevitably lose some global information useful for an extraction task.
In particular, only relying on local node features can cause failure when value nodes have no obvious patterns themselves or their local features are very similar to other non-value nodes.
To mimic the signal that may be available through visual features used in rendering-based methods, 
we use a relational neural network as our \textit{second module} (Sec.~\ref{ssec:pair_module}).
This allows us to model the relationship between a pair of elements using both distance-based and semantic features.
The rationale behind this is to learn more global representations of node pairs so that we can jointly predict node labels instead of relying only on local features.

Extensive experimental results on a large-scale public dataset, the Structured Web Data Extraction (SWDE) corpus~\cite{Hao2011FromOT}, show that our model consistently outperforms competitive baseline methods by a large margin. 
The proposed \textsc{FreeDOM} is able to generalize to unseen sites after training on a small number of seed sites. In fact, we show that with training data from just three seed sites, our approach out-performs techniques that use explicit visual rendering features by 3.7 F1 points on average. To the best of our knowledge, our framework is among the first neural architectures that efficiently obtains high-quality representations of web documents for structured information extraction. 

\eat{Our framework utilizes minimal human efforts in feature engineering and does not require any rendering results, thus making large-scale information extraction on web documents much easier and more effort-light.
We believe the proposed model can be promising in other applications that require neural representations of web documents.}

%% file: sec_2_problem.tex
\section{Problem Overview}
\label{sec:problem}

In this section, we describe the extraction problem formally before outlining the solution architecture.

\smallskip
\subsection{Problem Formulation}

Given a \textbf{vertical} $v$ (e.g. \textit{Auto}) and a set of websites $W_v$ in this vertical, 
we assume that only a few \textbf{seed websites} $W_v^{s}$ are labeled by human annotators, 
and the other sites are unseen and unlabeled \textbf{target websites} $W_v^{t}$. 
A website $w_i$ can be viewed as a collection of web pages {$w_i=\{p_j\}$} that share similar layout structures.
Each page $p_j$ describes a single topic entity (e.g. a car model in \textit{Auto} vertical) and is parsed to a DOM tree $t_j$.

We formulate the extraction task  as \textit{node tagging} on DOM Trees.
For instance, the HTML page in Figure~\ref{fig:intro} can be parsed into a DOM tree, and each element on this page that we can view in the browser corresponds to a unique XPath and a DOM node.
Structured information extraction can be viewed as classifying each node into one of the pre-defined vertical-specific groups: 
\textit{``Model''}, \textit{``Engine''}, \textit{``MSRP''}, \textit{``Fuel\_Economy''}, and \textit{None}. 
Note that we only need to tag leaf nodes that have textual content.
Following the convention in structured web information extraction~\cite{Hao2011FromOT}, we assume a single leaf node can correspond to at most one target field.

The research question now is to learn a model $\mathcal{M}$ that is trained with labeled pages in a few seed sites $W_v^{s}$, but still works well on unseen target sites $W_v^{t}$ in the same vertical $v$, even though their templates may be completely different.
In other words, we want to learn site-invariant feature representations such that the model generalizes to unseen sites without additional human effort.




%% file: sec_3_overview.tex
\subsection{Model Architecture}
\label{sec:model}
We solve this problem using a two-stage architecture.
The first stage learns a dense representation for each DOM node by combining both markup and textual content from its neighborhood. 
We find that using both markup and text can result in better embeddings. 
This dense representation is used to train a classifier to predict if the node corresponds to a target field or is \textit{None}.


\begin{figure}[t]
	\centering
	\includegraphics[width=0.9\linewidth]{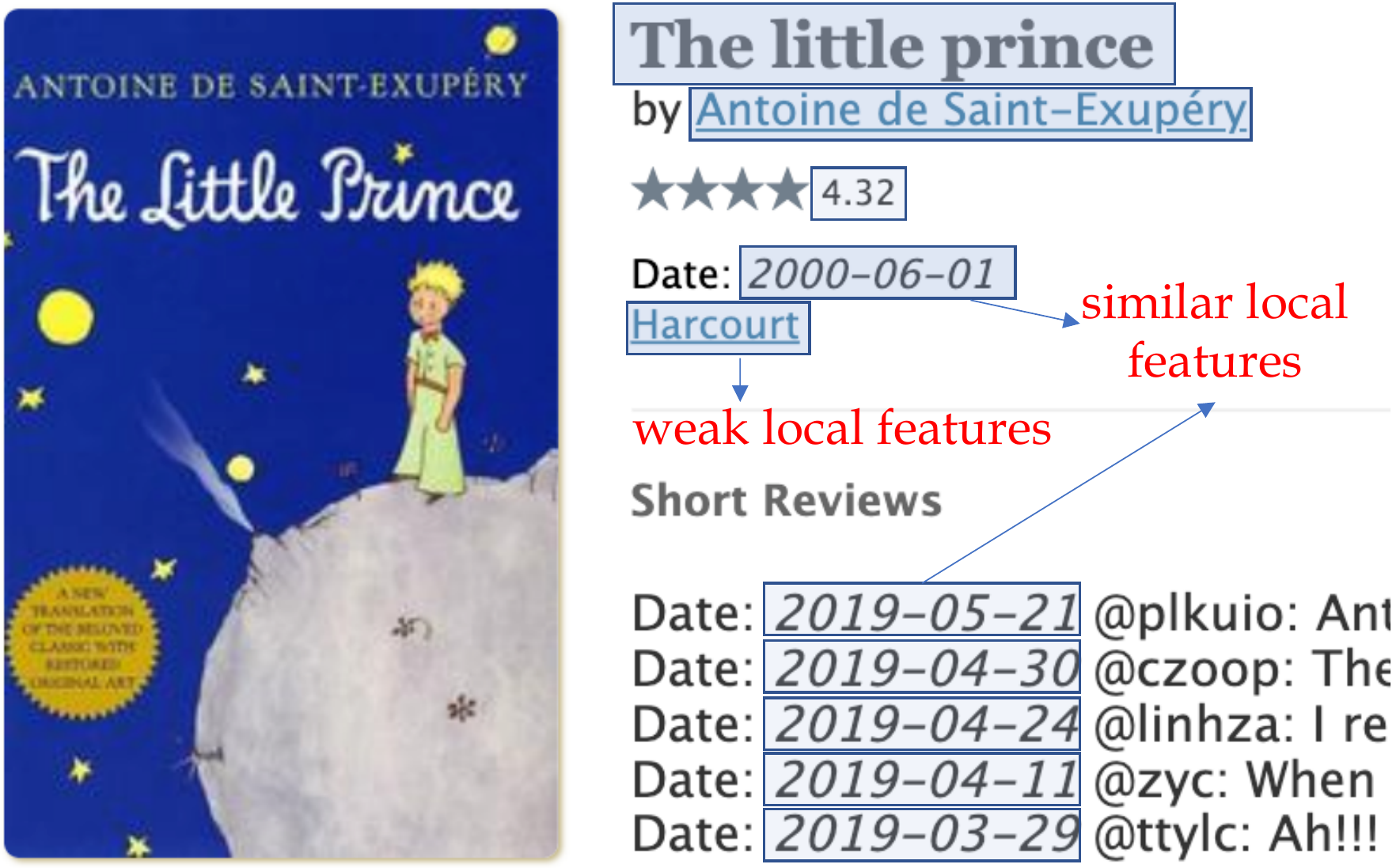}
	\caption{
	{The two reasons why we need to model structured dependency via relational features. 1) The node ``Harcourt'' should be predicted as \texttt{Publisher}, but was classified as \textit{None} by node-level classifier because it does not have very strong local features. 2) The top \texttt{PublishDate} values ``2000-06-01'' and the bottom \textit{review} dates ``2019-xx-xx'' share very similar local features, and thus the node-level model tends to predict all as \textit{None} because the majority have \textit{None} labels.}	}
	\Description{}
	\label{fig:nodeeg}
	\vspace{-11pt}
\end{figure}

We find that directly using the outputs from the first stage (i.e. node-level classifier) has a common failure:
the model fails to find some target fields and ends up labeling them \textit{None}. For these fields, the ground truth node either often shares very similar local features as other \textit{None} nodes or has very weak local signal (e.g. lacking of important anchor words). See the examples of \texttt{Publisher} and \texttt{PublishDate} in Fig.~\ref{fig:nodeeg}.

In retrospect, these problems were not surprising since the node representations captures local information rather than long-range patterns and constraints in the DOM tree.
Both cases are easy for humans to recognize because we can infer by their relations with other fields such as \textit{Author} and \textit{Title}.
In order to overcome these problems, we use a second stage node-pair model to incorporate information from parts of the document that are farther away.
The core idea was inspired by the relation network module~\cite{Santoro2017ASN} for modeling pairs of spots in images.
The node pair model is trained on pairs of DOM nodes and predicts if each of the nodes in the pair contains \emph{some} value.
The predictions of the second stage allows us to carefully discard low-confidence predictions from the first stage.

\begin{figure}[t]
	\centering
	\includegraphics[width=0.8\linewidth]{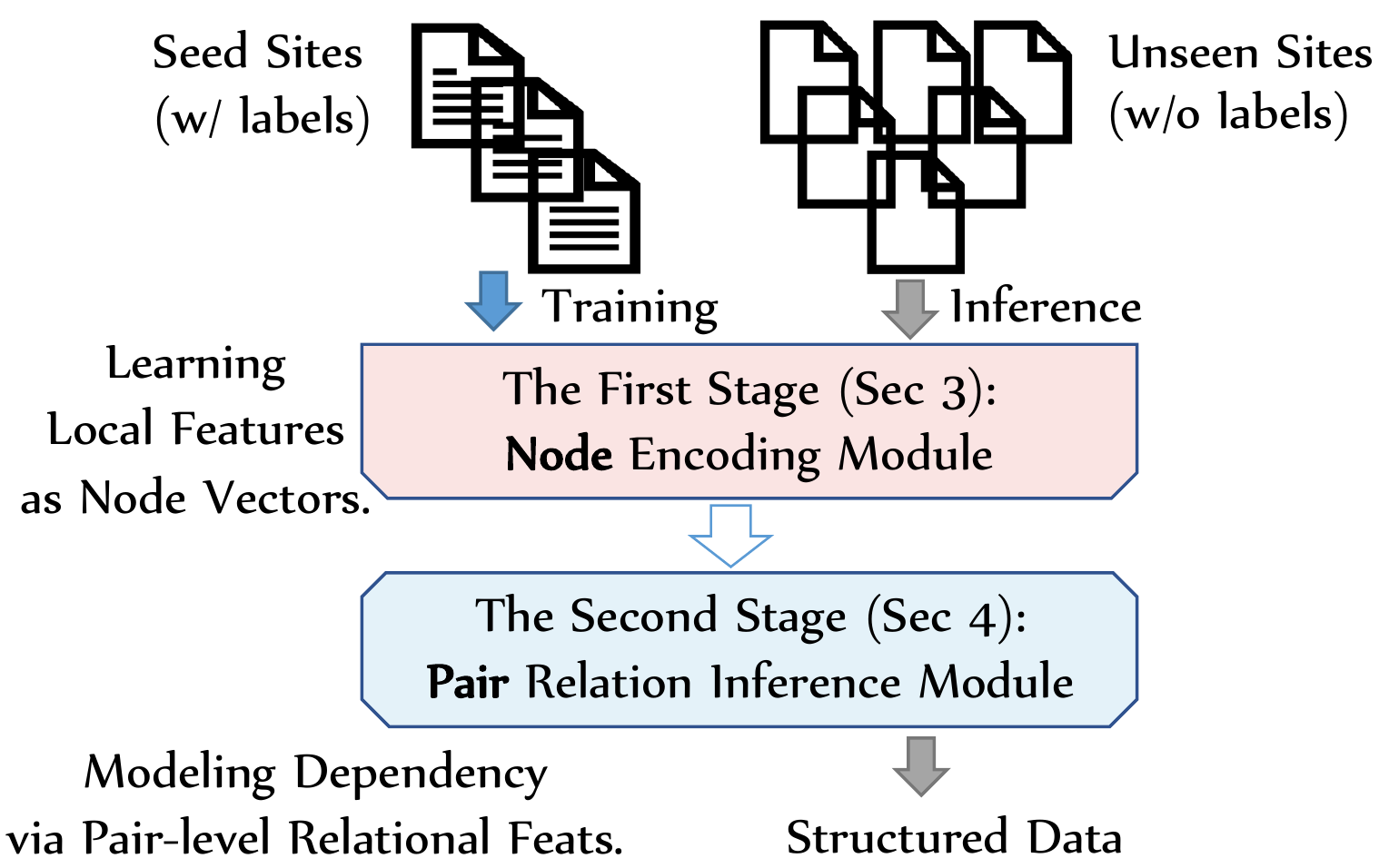}
	\caption{The overall workflow  of \textsc{FreeDom} .}
	\label{fig:overview}
\end{figure}
One question that immediately comes to mind is why not build on a sequence labeling model like BiLSTM-CRF~\cite{Lample2016NeuralAF} that is known to perform well on structured prediction tasks.
In Section~\ref{ssec:results}, we tackled this question and found that sequence tagging models built on top of the node embeddings produced by the first stage work slightly worse than simply using the node-level classifier. 
However, the pair-level module is able to better encode long-range patterns and outperforms these alternatives. See Section~\ref{sec:exp} for experiments and analysis comparing with a BiLSTM-CRF.

%% file: sec_4_node.tex
\section{Node Encoding Module}
\label{ssec:node_module}
We describe the first stage of the \textsc{FreeDOM} architecture which solves a node-level classification task for each ``variable''\footnote{Variable nodes (with the same XPath) have different contents across different pages (as shown in blue boxes in Figure~\ref{fig:nodeeg}). Thus, we can ignore nodes that are common boilerplate, such as navigation bars, headers, footers, etc. See~\ref{ssec:vni} for more details.} leaf node in a DOM tree.
We learn a dense representation for each node by combining several features associated with each node.


\subsection{Input Features}
\label{ssec:nodefeat}
\begin{figure}[t]
	\centering
	\includegraphics[width=1.01\linewidth]{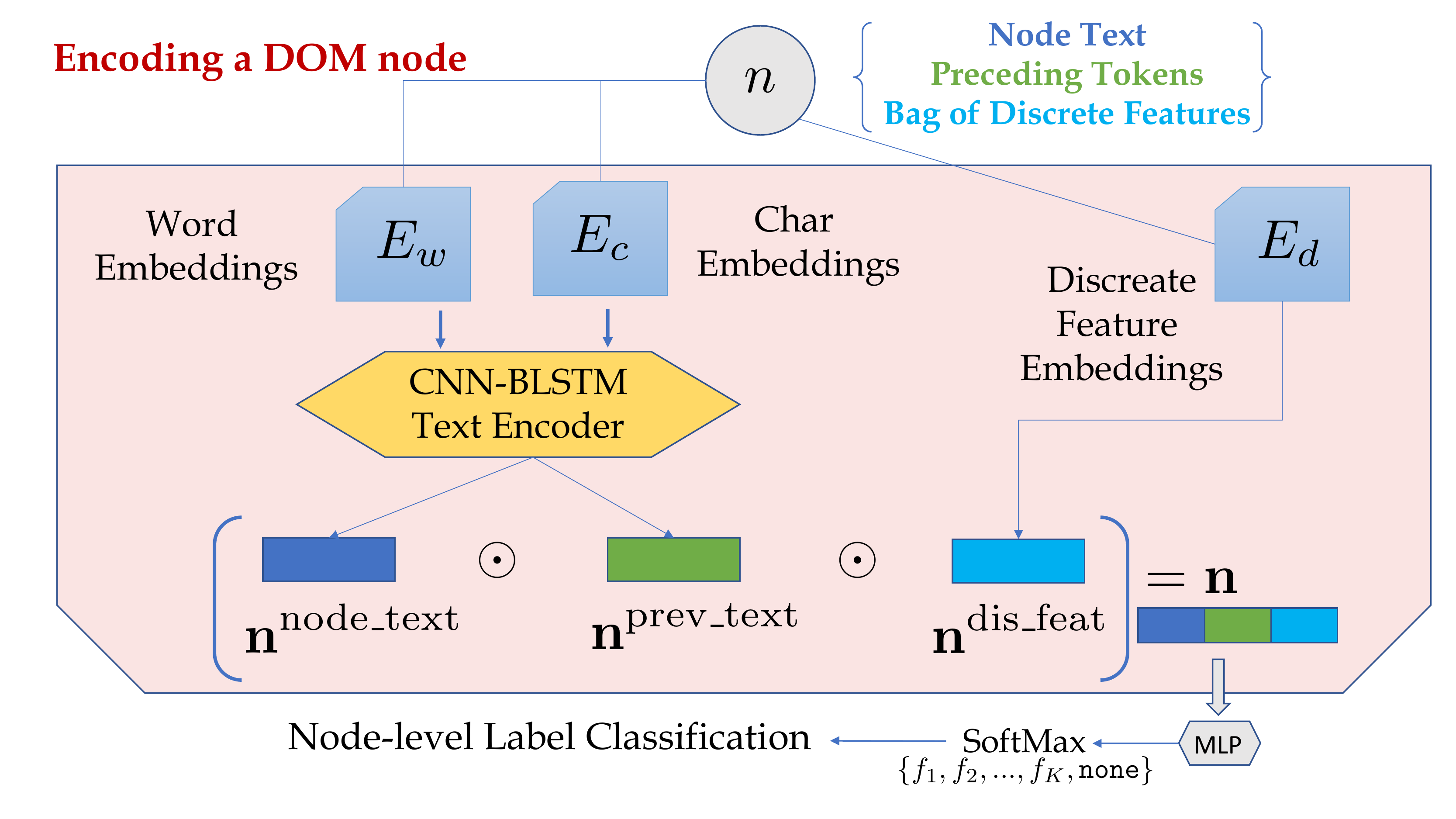}
	\caption{Learning a node-level classifier with  trainable comprehensive node embeddings from three different views. \vspace{-10pt} }
	\label{fig:nodemodel}
\end{figure}

We consider three primary sources of information to learn a representation for each node: 
1) node text (i.e., text tokens in the node), 
2) preceding text, and
3) discrete features based on markup and type annotations.
As shown in Figure~\ref{fig:nodemodel}, 
the node-level module first encodes each of these node-level features individually with separate networks.
These embeddings are concatenated and we train a simple feed-forward network over this representation via optimizing the SoftMax classification loss.

\smallskip
\noindent
\textbf{{Node text embeddings.}}\quad 
The text inside a leaf DOM node $n$ can be seen as a sequence of tokens $\{w_i\}$.
Each token itself is a sequence of characters $\{c_{j}\}_i$, where $w_i \in \mathcal{W}$ and $c_j \in \mathcal{C}$. 
We use a two-level neural sequence encoder to capture both sub-word and word-level semantics.
Modeling char-level semantics is beneficial for modeling the \textit{morphological patterns} in values of different fields.
General patterns consisting of out-of-vocabulary words and special characters can be learned with char-level CNNs and word-level LSTMs without explicit expert knowledge~\cite{Hao2011FromOT}. 
For example, a node for the ``\texttt{Fuel\_Economy}'' field in \textit{Auto} vertical could be ``city 33 hwy 27'', and capturing such ``city \textit{xx} hwy \textit{yy}'' patterns requires both token-level and char-level representations.

Similar to the architecture used for named entity recognition~\cite{ma2016end},
we first initialize a character embedding look-up table $\mathbf{E}_c \in \mathbb{R}^{|\mathcal{C}| \times \texttt{dim}_c}$ for all characters ($\mathcal{C}$)
where $\texttt{dim}_c$ is a hyper-parameter for the dimension of character embedding vectors.
Then, we use 
a convolutional neural network (CNN) to efficiently encode the sequences of character embedding vectors. 
After pooling by the CNN, we have a character-level word representation, which is further concatenated with a word-level vector initialized using external Stanford-GloVE~\cite{pennington2014glove} word embeddings.
On the top of such concatenated word representations,
we employ a bidirectional LSTM network for contextually encoding the whole sequence forward and backward.
The final node text representations $\mathbf{n^{\text{node\_text}}}$ can be derived as follows: 
\begin{gather*}
	\mathbf{c^{i}} = \texttt{CNN}(\{c_1, c_2, ..., c_{|w|}\}) ~;~
	\mathbf{t_i} = [ \mathbf{w^{i}} \odot \mathbf{c^{i}}] \\
	\mathbf{n^{\text{node\_text}}}  = \texttt{AVG}[\texttt{LSTM}_f(\{\mathbf{t_1}, \mathbf{t_2}, ...\}) \odot \texttt{LSTM}_b(\{\mathbf{t_{|n|}}, \mathbf{t_{|n|-1}}, ...\}) ]
\end{gather*}
where $[\cdot \odot \cdot ]$ denotes concatenation. 

~\\
\noindent
\textbf{{Preceding text embeddings.}}
{The text in the nodes that precede the target node is clearly a valuable signal.}
Suppose that we are extracting the ``MSRP'' nodes on a page, and the ground truth is the node with text ``\$9,970''.
There may be many other nodes in the page also containing similar texts (e.g. a node with ``\$9,872') denoting the invoice price instead of the target MSRP field.
The text preceding the target node, with tokens like ``MSRP :'' and ``Invoice Price :'' in our example, is clearly a critical signal.


We use the same text-encoding module (i.e. the CNN-BLSTM network) to encode the content preceding the target node (i.e. the 10 previous tokens).
This results in an additional vector representing the previous node text as $\mathbf{n^{\text{prev\_text}}}$.

~\noindent \textbf{{Discrete features.~~}}
The third source of information comes from markup and type-specific annotations on the content in the target node. We model these as a bag of discrete features.
The {DOM leaf type} of the leaf node such as `<h1>', `<div>', `<li>', `<span>' can be useful for identifying some fields. For instance, key fields like like model names and book titles are often encoded using <h1>.
We also apply a set of straightforward \textit{string type checkers} (e.g. those in the NLTK toolkit) to know if the node text contains any \textit{numbers}, \textit{dates}, \textit{zip codes}, \textit{URL link}, etc.
Note that a single node can have multiple such discrete features of these two groups.
We model these as follows:
we first maintain a look-up embedding table $\mathbf{E_{d}} \in \mathbb{R}^{|\mathcal{D}|\times \texttt{dim}_d}$, where $\mathcal{D}$ is the set of all possible discrete features and $\texttt{dim}_d$ is the dimension of feature vectors.
We utilize max pooling to map two multi-hot discrete feature vectors respectively to each bag of features $\mathbf{d_1}, \mathbf{d_2} \in \{0, 1\}^{|\mathcal{D}|}$ to real space vectors of dimension $\texttt{dim}_{d}$: $\mathbf{n}^{\text{dis\_feat}} = [ \mathbf{d_1} \mathbf{E_d} \odot \mathbf{d_2} \mathbf{E_d}]$ by matrix multiplication.
This can be seen as the concatenation of two max-pooled vectors over the associated feature-vectors in each bag of discrete features.
The three sources of information described are each encoded into dense vectors. These three vectors are concatenated to provide a comprehensive representation of the node in the DOM tree.


\subsection{Learning Node Vectors via Classification}
\label{ssec:nodeft}
The concatenated output representation is connected to a multi-layer perceptons (MLPs) for multi-class classification via a \texttt{SoftMax} classifier (see Figure~\ref{fig:nodemodel}) that  is illustrated in the following equations:

\begin{gather*}
\mathbf{n} = \left[ \mathbf{n}^{\text{node\_text}} \odot \mathbf{n}^{\text{prev\_text}} \odot \mathbf{n}^{\text{dis\_feat}} \right]\\
\mathbf{h}=\texttt{MLP}(\mathbf{n}), ~ \text{where}~ \mathbf{h} \in \mathbb{R}^{K+1} ~;~
l = \arg\max_{i} \frac{e^{\mathbf{h}_i}}{\sum_{j=1}^{K+1}e^{\mathbf{h}_j}}
\end{gather*}

Recall that our goal is to assign a label $l$ for each node $n$ in any page $p$, where $l$ is either a pre-defined field (out of $K$ types) or a ``\texttt{none}'' label.
Thus, the outputs from the SoftMax represent the distribution over the target labels, and the intermediate output over which the MLP is the learned combined representation for each node.
The outputs from the first stage can directly be used as (tentative) labels for each node.

%% file: sec_5_relation.tex
\begin{figure}[t]
	\centering
	\includegraphics[width=1.\linewidth]{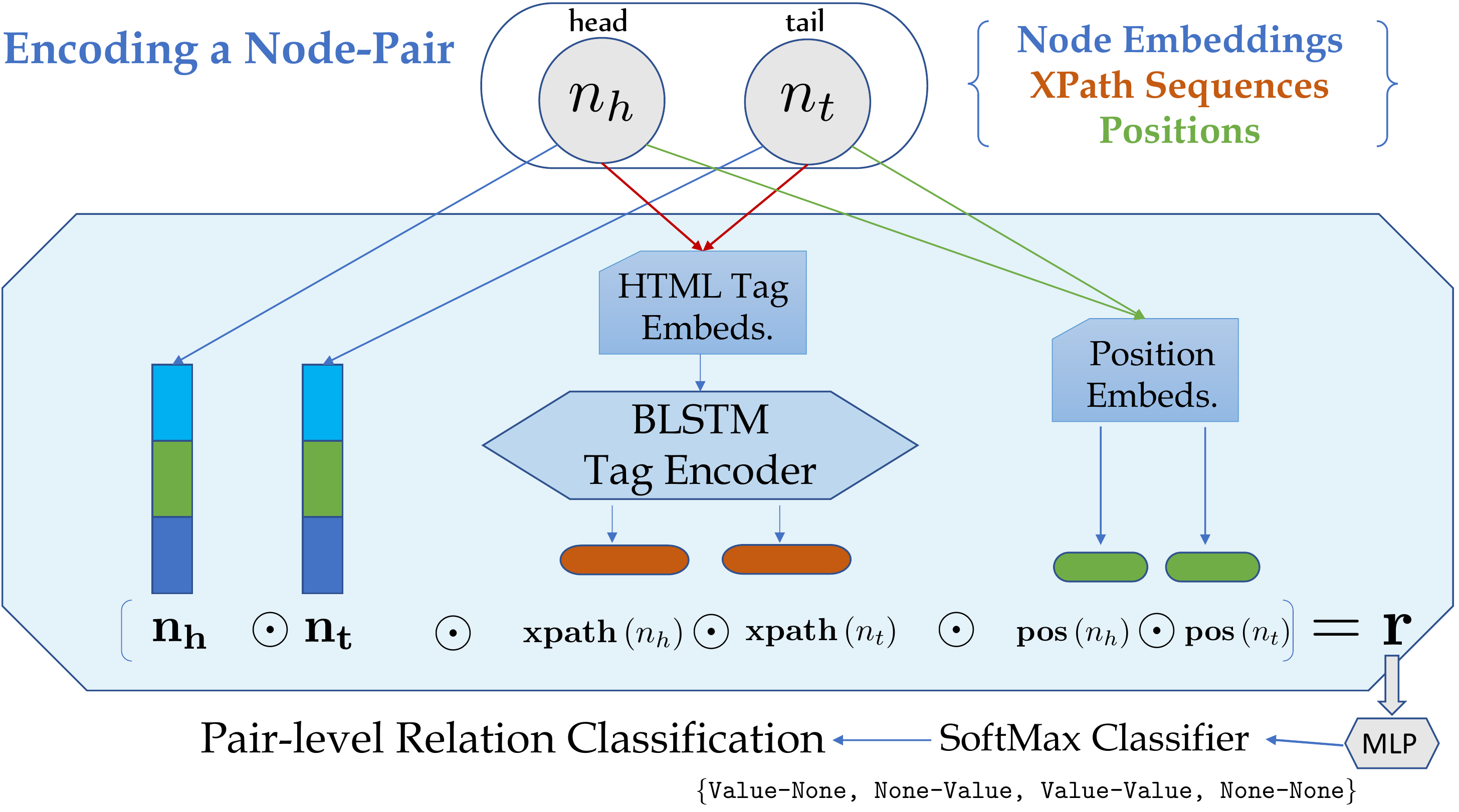}
	\caption{{Learning node-pair representations from different signals and the pair-level classification model. }}
	\vspace{-10pt}
	\label{fig:pairmodel}
\end{figure}

\section{{Relation Inference Module}}
\label{ssec:pair_module}

As we previously mentioned, directly using the outputs from the first stage may end up not predicting a required field for any of the nodes in the page. Recall the example in Figure~\ref{fig:nodeeg},
the first stage model assigns a 'None' to the \texttt{PublishDate} field node since it has very similar features (e.g. ``Date'' as preceding text) as many 'None' nodes.
Missing the prediction of the \texttt{Publisher} field is instead because its local features are not strong enough to generalize to unseen sites.
These issues are expected given the first stage model does not use non-local information and does not take into account global page-level constraints. 
Individually classifying node labels inherently limits the model's ability of using a broader context for jointly classifying labels of multiple nodes within a page.
Instead, we argue that modeling conditional dependencies among the node-level neural features can help improve the output predictions.


Knowing that value nodes of related fields are generally presented in a similar layout across sites can be valuable in distinguishing between these similar nodes. 
Figure~\ref{fig:distviz} in Appendix~\ref{sec:appendix} shows that pairwise distances between target fields are correlated across sites.
Rendering-based models usually design heuristic methods to use visual distance to deal with these problems in the post-processing stage, which are more expensive and need human knowledge.
The success of visual distance features used in rendering based models also supports the claim that such relations between fields are strong transferable patterns~\cite{Hao2011FromOT}.

\textit{\textbf{How can we model such distance patterns in a neural network for learning end-to-end instead of hard-coding as rules?}}
We propose to do this in the second stage of our architecture to capture conditional dependencies between nodes when predicting node labels.
However, modeling the dependencies among a huge number of nodes in a DOM tree structure is challenging.
Inspired by the recently proposed relation network module~\cite{Santoro2017ASN} for modeling pairs of spots in images, 
we propose to directly model the relation features of node-pairs on DOM trees.
This should allow us to efficiently encode relational features between them.
With such relational features, 
the conditional dependencies can be used to produce more precise structured predictions.

We first propose a way to construct node pairs (Sec.~\ref{ssec:paircons}) based on the first-stage outcomes, then we learn a node-pair relation network (Sec.~\ref{ssec:pair_init}) for encoding node-pairs with additional distance-based features, and finally we show how to aggregate the node-pair predictions to improve the predictions from the first stage for structured data extraction (Sec.~\ref{ssec:aggvot}).

\subsection{{Node-Pair construction}}
\label{ssec:paircons}
A typical web page has over 300 nodes (see Tab.~\ref{tab:stat}) that our model should predict labels for.
Considering that the full combinations (about $300^2$ node-pairs) is huge, 
we instead only consider the following subsets to reduce the
computational cost in the second stage.
Based on the first-stage outputs,
we use the raw output scores ($\mathbf{h_i}$) to filter the nodes as follows.
First, we divide the fields to two groups: \textbf{certain fields }and \textbf{uncertain fields} depending on whether there is at least one node predicted for it.
For the certain fields, we take each of the nodes predicted by the first-stage module as the anchor nodes for the field.
While for uncertain fields,  we instead take the top $m$ nodes with the highest score for that field to construct node-pairs.
Say we have $K$ fields in total, and $T$ of them are \textit{certain fields}, then we have $T*(T-1) + 2*T*(K-T)*m + (K-T)*(K-T-1)*m^2$ node-pairs\footnote{Note that $2*T*(K-T)$ stands for the number of  field pairs containing one certain and one uncertain fields and $*m$ means that we iterate the top $m$ nodes in uncertain fields to construct pairs.
Similarly, $(K-T)*(K-T-1)$ is the number of pairs where both fields are uncertain. $T*(T-1)$ is the number of certain fields pairs.} to score. 
For each node-pair, we assign a pair-label according to whether head/tail node is a value/none. 
That is,
we use the label space as \{(\texttt{N,N}),  (\texttt{N,V}), (\texttt{V,N}), (\texttt{V,V})\}, where \texttt{N}=``None'' and \texttt{V}=``Value''.
Thus, the learning objective of the second-stage is to predict the pair-labels $(l_{\texttt{\texttt{head}}}, l_{\texttt{\texttt{tail}}})$ of each node-pair $(n_{\texttt{\texttt{head}}}, n_{\texttt{\texttt{tail}}})$.
The \textit{key rationale} behind our pair construction process is that we hope the certain fields can work as pivots for the model to extract confusing fields by their distance-based features.

\subsection{Node-pair relation networks}
\label{ssec:pair_init}

\eat{ 
We argue that an evident relational feature is the distance between head node and tail node in each node-pairs in the page.
As shown in Figure~\ref{fig:distviz},
we find that different websites share similar patterns in the distance between value nodes of different fields\footnote{The distance values in the figure are the number of variable nodes are there between two fields which are further normalized by the total number of nodes.}.
The success of visual distance features used in rendering based models also support the claim that such relations between fields are strong transferable patterns~\cite{Hao2011FromOT}.
How can we model such distance patterns in a neural network for learning end-to-end instead of hard-coding as rules?
}

We propose two ways of modeling distances between nodes: XPath sequences and positional embeddings.
We combine both to learn an encoding for the node-pair distance and concatenate it with the first-stage output (i.e. node vectors) as the representation of node-pairs for learning.
Denoting the \textit{head} and \textit{tail} nodes as $n_\text{h}$ and $n_\text{t}$ respectively, we summarize the three views of node-pair representations in Figure ~\ref{fig:pairmodel}).

An XPath of a DOM node can be seen as a sequence of HTML tags like [``<html>'', ``<body>'', ``<div>'', ``<ul>'', ``<li>''].
{Nodes that are closer in the rendered result by web browsers usually have also similar XPath sequences.}
The distance in their XPaths can thus be used as a proxy for estimating their distances in rendered pages by web browsers.
However, the underlying distance between a pair of XPaths is hard to model directly for generalizing to unseen websites.
Inspired by neural sequence modeling, if we could represent each XPath as a vector and then their hidden differences can be further modeled by the non-linearity of neural networks.
Specifically, we train a bi-directional LSTM (BiLSTM) to get the representation of an XPath sequence, such that the vectors of all tags in it is encoded as a dense vector.
We denote the XPath LSTM output vector of head and tail node as $\textbf{xpath}(n_h)$ and $\textbf{xpath}(n_t)$ respectively.

The position of a node in the page is also an explicit signal to describe the distance.
Thus, we further utilize positional embeddings for encoding, which is commonly used in relation classification tasks~\cite{Zeng2014RelationCV}.
We take the positions of each node in a limited range of integers between 0 and $L$ (say $L=100$).
Then, we will have an embedding matrix $E_{\texttt{pos}}$ to learn, and thus we have $\textbf{pos}(n_h)$ and $\textbf{pos}(n_t)$ for positional features of the two nodes.

From the first-stage model-level predictions,
each node has been represented as a node vector $\mathbf{n}$. 
We continue using such learned representation of all nodes, denoted as $\mathbf{n}_{\texttt{h}}$ and $\mathbf{n}_{\texttt{t}}$, which respectively represent the local features of the two nodes by the first-stage model. 
Finally, we concatenate all the three views of relational node-pair features together and have: 
$$\mathbf{r} = \left[
\mathbf{n}_{\texttt{h}}\odot\mathbf{n}_{\texttt{t}}\odot \textbf{xpath}(n_h)\odot\textbf{xpath}(n_t) \odot
 \textbf{pos}(n_h)\odot\textbf{pos}(n_t) \right]$$
On top of this, we further build a multi-layer feed-forward neural networks as a SoftMax multi-class classifier.
Unlike what we do in the node-level module, here our target label space for node-pairs are \{\texttt{none-none}, \texttt{none-value}, \texttt{value-none}, \texttt{value-value}\}.

\subsection{Label-pair aggregation and voting.}
\label{ssec:aggvot}
\textbf{Pair-labels aggregation.}\quad Recall that in the pair construction stage (Sec.~\ref{ssec:paircons}), we choose one node for each certain field (out of $T$) and $m$ nodes for each uncertain field (out of $K-T$).
Thus, each node in the node-pairs is already assigned a field-type.
We only need to re-consider candidate nodes for uncertain fields and determine if they should be treated as a
\texttt{Value} or a \texttt{None} for the target field.
The prediction of the second-stage model for a node pair assigns a \texttt{Value} or a \texttt{None} to nodes in the pair. A node involved in $X$ pairs then gets $X$ labels. It is considered as \texttt{Value} if >= $N$ labels among the $X$ labels are \texttt{Value}, where $N$ is a hyper-parameter. 
This is the output from the second stage.


\smallskip
\noindent
\textbf{Site-level XPath Voting.}\quad 
Inspired by prior works on web content redundancy~\cite{Gulhane2010ExploitingCR, Li2012ExploitingAR}, 
we argue that many value nodes of a field tends to come from the same XPath across different pages in the same site. 
Thus we find the XPath selected as the field value by the majority pages and correct the rest of the pages (if there is inconsistent outliers) to extract field value from this XPath as well.

%% file: sec_6_exp.tex
\section{Experiments} 
\label{sec:exp}
The experiments in the section are designed to answer two broad questions. First, how well does \textsc{FreeDom} generalize to unseen sites in a given vertical after being trained on examples from a small number of seed sites? We present a comparison with previous methods proposed in the literature. We show that \textsc{FreeDom} can generalize better to unseen sites after training on just three seed sites and beats the previous SOTA by 3.7 points \emph{without using visual renderings}. Second, how much do various design choices contribute to our accuracy? We present an ablation study including a comparison to sequence labeling baselines (e.g. BLSTM-CRF) showing the contributions of the two modules in the solution.

\eat{
We first introduce the dataset and experiment setting~(Sec.~\ref{ssec:dataset}), then present the baseline methods~(Sec.~\ref{ssec:baseline}), and finally compare the performance to evaluate our proposed model~(Sec.~\ref{ssec:results}) with detailed analysis and discussion.
}

\begin{figure*}
	\centering
	\includegraphics[width=1.0\linewidth]{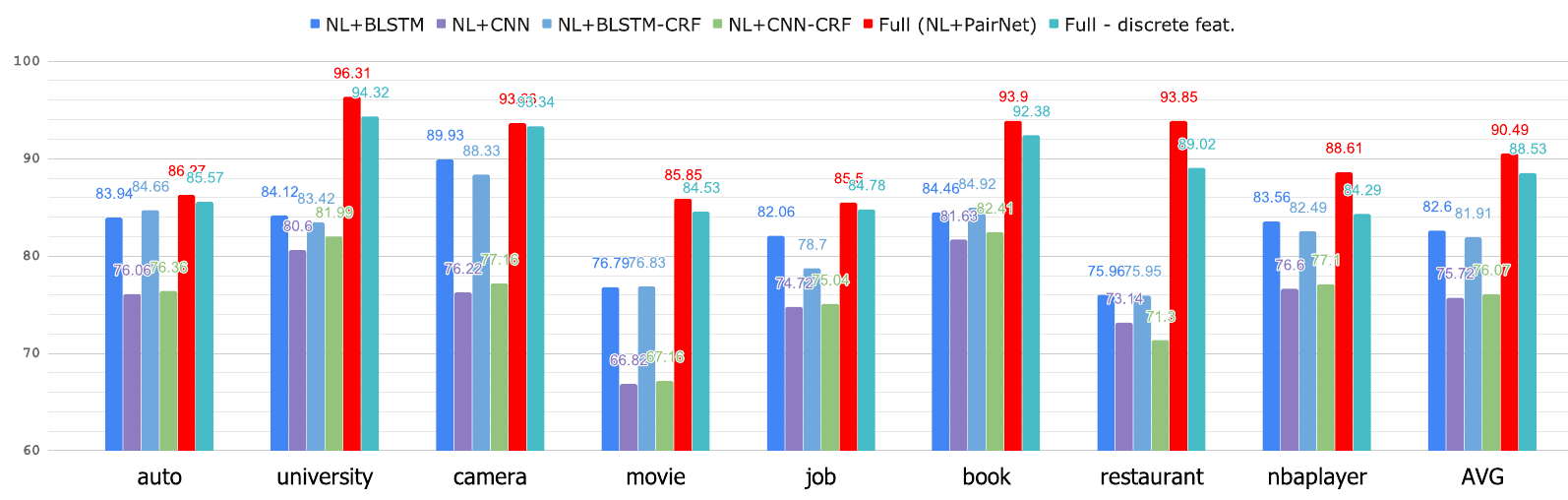}
	\vspace{-2.5em}
	\caption{Ablation study results by replacing the proposed Pair Network with other node-sequence labeling-based models (CNN/BLSTM + an optional CRF layer) when using three seed websites in each vertical ($k=3$).  We also show the effectiveness of Freedom-Full if discrete features are disabled (w/o discrete feat.). }
	\label{fig:abalation}
\end{figure*}
\subsection{Dataset and Set-up}
\label{ssec:dataset}
~\\
We use the publicly accessible \textit{Structured Web Data Extraction} (SWDE) dataset~\cite{Hao2011FromOT} for all our evaluation tasks.
As shown in Tab.~\ref{tab:stat}, the SWDE dataset has 8 verticals and 10 websites for each vertical. 
Each vertical specifies 3$\sim$5 fields to extract. 
Each website has hundreds of pages, where a page has about 300 variable nodes for us to classify.
We use open-source LXML library\footnote{\url{https://lxml.de/} (Other alternatives can also work.)} for efficiently processing each page to get its DOM tree structure.

\begin{table}[t]
	\scalebox{0.72
	}{
	\begin{tabular}{@{}cccccl@{}}
		\toprule
		\textbf{Vertical}   & \textbf{\#Sites} & \textbf{\#Pages} & \textbf{\#Var. Nodes} &   \multicolumn{1}{c}{\textbf{Fields}}               \\ \midrule
		\textbf{Auto}       & 10               & 17,923    & 130.1                          & model, price, engine, fuel\_economy                   \\
		\textbf{Book}       & 10               & 20,000    & 476.8                          & title, author, isbn, pub, date \\
		\textbf{Camera}     & 10               & 5,258      & 351.8                          & model, price, manufacturer                            \\
		\textbf{Job}        & 10               & 20,000     & 374.7                            & title, company, location, date\_posted                \\
		\textbf{Movie}      & 10               & 20,000    & 284.6                          & title, director, genre, mpaa\_rating                  \\
		\textbf{NBA Player} & 10               & 4,405    & 321.5                             & name, team, height, weight                            \\
		\textbf{Restaurant} & 10               & 20,000   & 267.4                            & name, address, phone, cuisine                         \\
		\textbf{University} & 10               & 16,705    &  186.2                           & name, phone, website, type                            \\ \bottomrule
		\vspace{0pt}
	\end{tabular}
	}
	\caption{The statistics of the public SWDE dataset. \vspace{-2.5em}}
	
	\label{tab:stat}
\end{table}

\textbf{Experimental settings.}\quad
We randomly select $k$ seed websites as the training data to train different extraction models, and test the models on the pages in other remaining $10-k$ sites (i.e. the target sites) where \textbf{\textit{no label at all is used for transferring trained models}}.
For each vertical, we take 10 cyclic permutations after fixing an order within the websites.
Obviously, the same website is never present in training and test data in any given experiment.
This setting is close to the real application scenarios we care about where training data is generated from a small number of seed sites, and we want to learn to extract for dozens of other unseen sites where we may have no labels. 
\quad 
We use the same evaluation metrics following the authors of SWDE~\cite{Hao2011FromOT} and compute the page-level F1-scores in each vertical and field.
Note that the authors of SWDE do not consider the cases where there are multiple values for the same field in a page and thus we take our top-1 predictions. 
 
\subsection{Baseline methods and FreeDOM variants}
\label{ssec:baseline}
The \textbf{Stacked Skews Model} (\texttt{SSM})~\cite{carlson2008bootstrapping} utilizes a set of human-crafted features for aligning the target unseen website DOM trees to the pages in existing annotated seed websites with 
tree alignment algorithms~\cite{Zhai2005WebDE} that are time-consuming.
The authors of the SWDE dataset applied this method under the same problem setting and stated that this was the state-of-the-art method that did \emph{not require} visual rendering features.
Thus, this model is the closest to ours in terms of assumptions on available supervision.

\begin{figure*}
	\centering
	\includegraphics[width=1.0\linewidth]{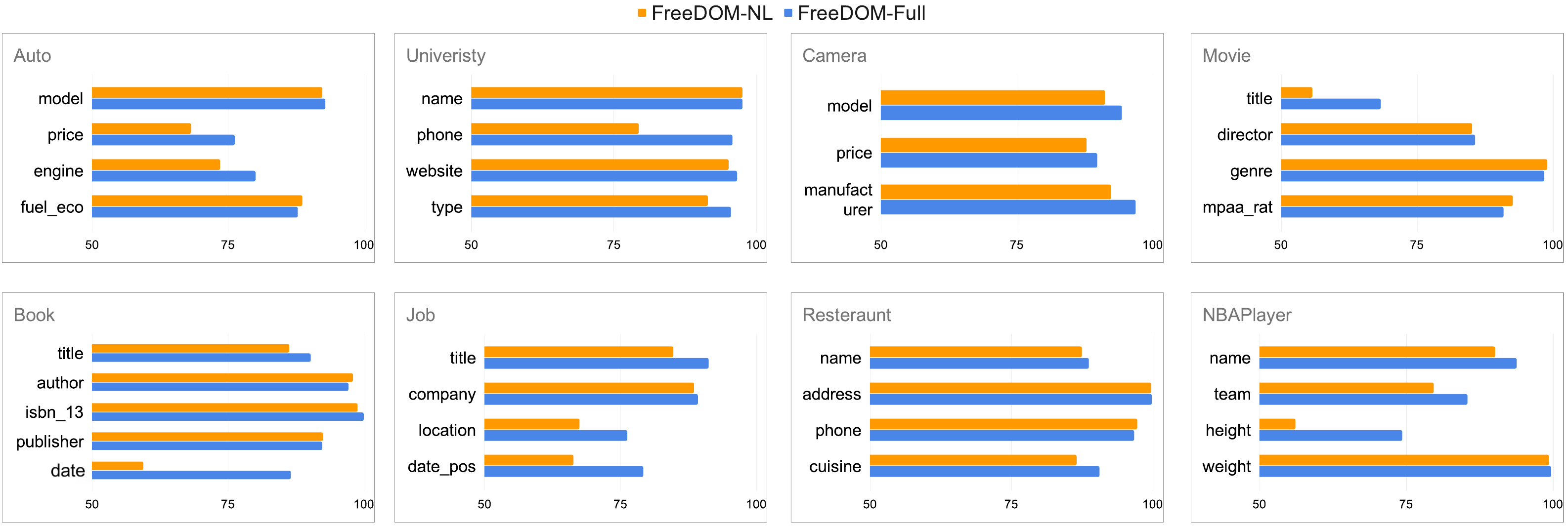}
	\vspace{-2em}
	\caption{Per-field performance (F1\%) comparisons between FreeDOM-NL and FreeDOM-Full when using three seed websites. }
	\label{fig:perfield}
\end{figure*}


Hao et al.~\cite{Hao2011FromOT} propose a set of \textbf{rendering-feature baseline methods} that use the visual features to explore the distance between each block in the web browser rendered result. 
The visual distances have proven to be very strong features for structured data extraction in web documents.
Note that the rendering requires downloading and executing a large amount of external scripts, style files, and images, which are extremely time/space-consuming.
Further, their approach requires a set of pre-defined, human-crafted patterns for recognizing values of fields (e.g. price, length, weight) using prefix and suffix n-grams.

{
This baseline method has three variants, which we named as \texttt{Render-PL}, \texttt{Render-IP}, and \texttt{Render-Full} respectively.
Hao et al. \cite{Hao2011FromOT} first use the visual-distance features and human-crafted rules to learn a model to classify each node individually at the page-level (\texttt{Render-PL}).
Using site-level voting like we mentioned in Sec.~\ref{ssec:aggvot}, they further improve the method by controlling inter-page consistency (\texttt{Render-IP}).
Finally, they further propose a complicated heuristic algorithm for computing visual distances between predicted value nodes and adjust the predictions in a whole site (``{Render-Full}'').
As we can see, these methods\footnote{We report the numbers exactly from the paper of such models~\cite{Hao2011FromOT} without re-implementation since some of their resources are not publicly available or outdated.} are highly dependent on the visual rendering to obtain strong distance features as well as hand-crafted patterns.

We also compare the proposed \textsc{FreeDom} model, named ``{FreeDOM-Full}'', with the node-level model (``{FreeDOM-NL}''), which simply uses the output for the first stage.
Comparing ``{FreeDOM-NL}'' and ``{FreeDOM-Full}'' can tell us the importance of our relation inference module (Sec.~\ref{ssec:pair_module}).
We also adopt a series of recently proposed neural sequence modeling architectures, as the alternatives to the our proposed pair-level networks.
These sequence labeling models are based on our node-level sequence encoder and an optional CRF~\cite{Lafferty2001ConditionalRF} layer to perform node tagging.
These are also reasonable baselines since they allow the model to learn from a longer inter-node dependency while performing structured predictions.
According to the choices of the node sequence encoder and whether to impose CRF tagging layer, 
there are \texttt{FreeDOM-NL+BLSTM}, \texttt{FreeDOM-NL+CNN}, \texttt{FreeDOM-NL+BLSTM-CRF}, and \texttt{FreeDOM-NL+CNN-CRF}.
Note that all of them are further improved with site-voting for fair comparisons.
The tuned parameters of these models are reported in Appendix~\ref{sec:appendix}.

}

\begin{table}[t]
	\begin{tabular}{cccccc}
		\toprule
		{\textit{\textbf{Model} \textbackslash ~\#Seed Sites}} &
		\textbf{$k=1$} & \textbf{$k=2$} & \textbf{$k=3$} & \textbf{$k=4$} & \textbf{$k=5$} \\\midrule
		\texttt{SSM}                                   & 63.00      & 64.50      & 69.20      & 71.90      & 74.10      \\
		\texttt{Render-Full}                           & \textbf{84.30}      & { 86.00}      & { 86.80}      & 88.40      & 88.60      \\
		\texttt{FreeDOM-NL}                            & 72.52      & 81.33      & {86.44}      & { 88.55}     & { 90.28}      \\
		\textbf{FreeDOM-Full}                          & { 82.32}      & \textbf{86.36}      & \textbf{90.49}      & \textbf{91.29}      & \textbf{92.56}   \\ \bottomrule  
		~\\
	\end{tabular}
	
	\caption{{Comparing performance (F1-score) of the four typical methods including our FreeDOM using different numbers of seed sites (from 1 to 5). 
	Each entry is the mean value on all 8 verticals and 10 permutations of seed websites, thus 80 experiments in total.  Note that {Render-X} methods utilize rendering results that require huge amount of external resources than {SSM} and {FreeDOM-X}. \vspace{-15pt}}}
	\label{tab:overall}
\end{table}

\subsection{Experimental results and discussion.}
\label{ssec:results}

We first compare the overall performance of all baseline models and our full model using different numbers of seed websites.
Then, we do a detailed ablation study on different verticals and fields to verify the effectiveness of our proposed two-stage framework. 

\subsubsection{Overall performance}
\label{sssec:overallresult}
~\\
As shown in Tab.~\ref{tab:overall},
we extensively test the overall performance of four methods on how well they generalize to unseen sites using different numbers of seed sites for training.
Note that each entry in this table is the mean of the F1 of 80 experiments: 8 verticals and 10 permutations of the $k$ seed websites. 
We can see that the full rendering-based  model (\texttt{Render-Full}) perform the best when $k$=1, but when $k\ge2$ the performance of our proposed full \texttt{FreeDOM} outperforms other methods by a large margin (nearly $4$ absolute F1 increment over \texttt{Render-Full} and $18$ over \texttt{SSM} when $k=5$).

As a side note\footnote{In their paper, only ``Render-Full'' has detailed results from $k$=1 to 5 and it is significantly better than the other two. Therefore, we mainly compare with Render-Full.}, the Render-PL's result is 70.63\% and Render-IP's is 82.01\% when $k=1$.
It shows that rendering-based features and rule-based patterns can be stronger than our proposed neural architecture when there is very little training data.
This is not surprising since these visual features were carefully crafted by humans
to capture patterns present in SWDE.
However, when using more than two seed websites, our neural architecture quickly
matches the performance obtained from hand-crafted visual features.
\texttt{FreeDOM-NL} achieves a similar performance with \texttt{Render-Full} ($86.44$ vs $86.80$) when using 3 seed websites.
With our novel pair-level networks (\texttt{FreeDOM-Full}) for learning relational features, the performance is further improved.
A detailed report about the performance of \texttt{FreeDOM-Full} on different verticals are shown in Table~\ref{tab:vertical}.
We can see that the absolute improvement on F1 increases all the way to using 5 seed websites with gradually diminishing improvements.


\begin{table}[t]
	\scalebox{0.85}{
	\begin{tabular}{cccccc}
		\toprule
		{\textit{\textbf{Vertical} \textbackslash ~\#Seed Sites}}   & \textbf{$k=1$} & \textbf{$k=2$} & \textbf{$k=3$} & \textbf{$k=4$} & \textbf{$k=5$} \\\midrule
		\texttt{Auto}       & 84.78            & 83.33             & 86.27             & 88.37             & 88.50             \\
		\texttt{University} & 89.18            & 92.73             & 96.31             & 94.90              & 97.88            \\
		\texttt{Camera}     & 86.85            & 89.24             & 93.66             & 94.52             & 95.01            \\
		\texttt{Movie}      & 77.35            & 82.43             & 85.85             & 87.60              & 88.87            \\
		\texttt{Job}        & 78.35            & 80.11             & 85.50              & 85.30              & 86.60             \\
		\texttt{Book}       & 87.25            & 92.45             & 93.90              & 95.45             & 95.47            \\
		\texttt{Restaurant} & 81.07            & 88.54             & 93.85             & 92.89             & 94.67            \\
		\texttt{Nbaplayer}  & 73.71            & 82.03             & 88.61             & 91.26             & 93.46            \\ \midrule 
		\textbf{AVG}        & \textbf{82.32} & \textbf{86.36} & \textbf{90.49} & \textbf{91.29} & \textbf{92.56}\\
		\bottomrule
~\\		
	\end{tabular}
	}
\caption{The performance (F1-score) of \texttt{FreeDOM-Full} of using different numbers of seed websites in different verticals.}
	\label{tab:vertical}
	\vspace{-25pt}
\end{table}

\subsubsection{Ablation Study.}
\label{ssec:ablation}
~\\
To examine the effectiveness of our second-stage model,
we replace the pair network with a set of popular neural sequence labeling models. These can be seen as alternative methods that can use global context for producing structured predictions.
To ensure a fair comparison, we use the same first-stage model to compute the node representations.
In the end, we also test if an additional layer can help decode tag sequences, which is commonly used in sequence tagging models~\cite{Lample2016NeuralAF,Lin2017MultichannelBM, Lin2018NeuralAL}.\quad
As shown in Figure~\ref{fig:abalation},
we find that such sequence labeling-based models are all worse comparing to using the proposed pair networks in our second stage learning (even slightly worse than only using the node-level classifier). This is the case with both CNNs and BLSTM for encoding the sequences. Using CRF makes them even worse.
This raises an interesting question: Why do neural sequence labeling models not work well with DOMTree node classification?

Recent research efforts show that The CNN/BLSTM-CRF models are powerful for encoding input sequences and decode them into tag sequences.
However, we argue that sequences of DOM nodes ordered by their natural occurrence in web pages are not suitable for sequence learning models.
The reasons are three-fold: 1) node sequences are usually too long, 2) structural dependencies are not sequential, and 3) the labels are too sparse.
There are thousands of nodes in a DOM tree, and even if we filter all fixed nodes, the node sequences are still longer ($>$100). 
Also, consecutive nodes in such sequences do not necessarily have label-wise relations. Such dependencies on labels usually exist in distant node pairs, such that CNN-based encoder loses a lot useful relational information.
CRF is especially designed for capturing consistent label-transition probability, which is not evident in our sparse label sequences.
Therefore, these sequence labeling models do poorly as alternatives, even with our node encoder.

Apart from that,
we also show the importance of discrete features in our model (the last item in each group of Figure~\ref{fig:abalation}).
We conclude that such discrete features are helpful while removing them would not harm the performance too much.
Therefore, if running such discrete feature extractors might be a problem of users, they can still have a stronger model than using rendering-based methods.

{
\subsubsection{Per-field Analysis} \quad
\label{ssec:case}  
Figure~\ref{fig:perfield} shows the detailed per-field performance comparison between \texttt{FreeDOM-NL} and \texttt{FreeDOM-Full} when $k=3$.
We conclude that \texttt{FreeDOM-NL} can already perform very well on many fields, which means our proposed first-stage module learns great node-level representations. 
However, for certain fields that FreeDOM-NL is very bad like the \texttt{{publication\_date}} in Book vertical, the \texttt{FreeDOM-Full} utilize the relational features to improve them very much, and in the end significantly improve the overall performance.
Similarly, we find that \texttt{FreeDOM-Full} classifies better in the key fields (e.g. \texttt{Title} in \textit{Movie}, and in \textit{Job}).
This is because such values usually do not have directly useful local features learned by node-level model, but they can be identified by our pair-level model with relational features and joint predictions.
}

%% file: sec_7_related.tex
\section{Related Work}
\label{sec:relatedwork}

Our work builds on research in wrapper induction in the data mining community and the contributions of neural models for information extraction tasks in the NLP community. 
We apply it in the context of modeling semi-structured web documents with neural networks.
Specifically, we aim to build a more lightweight yet transferable model by getting rid of expensive rendering-based features and complex human-crafted algorithms.

\smallskip
\textbf{Structured Data Extraction.}\quad
Structured data extraction from web documents has been studied extensively in supervised settings~\cite{Sleiman2013ASO, ferrara2014web, schulz2016practical, Azir2017WrapperAF}.
Early works~\cite{Gulhane2011WebscaleIE, Melnik2011DremelIA, Gentile2013WebSI, Crescenzi2001RoadRunnerTA} usually require a significant number of human-crafted rules or labels for inducing a wrapper (i.e. a program or model for extracting values of interested fields), which is only used for a particular web site.
These wrappers are usually brittle when testing on unseen websites in the same vertical, although they can have high precision and recall on the training sites.
\quad 
Some recent works propose methods that can adapt to new websites. 
Zhai et al.~\cite{Zhai2005WebDE} employed active learning methods that can find the most influential new examples in target websites for human annotators to label, and therefore adapting the existing wrappers to new sites in a cheaper way.
However, in practice, active-learning methods not only constantly require  human effort in building specialized annotation tools~\cite{lin-etal-2019-alpacatag} but also need humans to label samples for each new site of interest.
A recent work by Lockard et al.~\cite{ceres} attempted to use additional knowledge bases as distant supervision to automatically label some samples in the target websites and then learn a machine learning model on such noisy-labeled data.
A large and comprehensive knowledge base is not always accessible and available for every domain. Consequently, their methods do not apply for emerging domains without first requiring the human effort of building a knowledge base.
To address the problem in a more domain-general setting, we do not impose any human prior knowledge on the values, and focus on a purely unsupervised model adaptation learning scenario.

\smallskip
\textbf{Transferable Extraction Models.}\quad
Hao et al.~\cite{Hao2011FromOT} proposed a method that was based on visual distance features on a rendered web page. This achieved promising results on unseen websites without using any new human annotations. 
Following their work, there are a series of rendering-based extraction models proposed for many different settings.
However, 
these rendering-based methods need to download all the external files including CSS style files, javascripts and images such that they can render a page with browser engines to know page layouts~\cite{omari2016lossless}.
In addition, they have to design heuristic algorithms for computing human-crafted distance metrics~\cite{Zhu20052DCR, Cohen2015SemiSupervisedWW}.
These drawbacks together make large-scale extraction less practical and efficient.
Our method, \textsc{FreeDOM}, instead is totally based on the HTML content itself without any use of external resources or page renders. When using two or more seed websites to train, \textsc{FreeDOM} outperforms even the approach with expensive human-crafted features on visually-rendered pages.

\smallskip
\textbf{Neural Architectures for Information Extraction.}~
\label{ssec:neuralie}
Another advantage of our method is that it can work well without any human-crafted features or patterns.
We adopt state-of-the-art sequence modeling techniques~\cite{Yang2016HierarchicalAN, Lample2016NeuralAF} from IE and NLP to build our char-based word representations, XPath LSTMs, and position embeddings.
Our pair-level network is inspired by the relational reasoning networks~\cite{Santoro2017ASN}, which learns to induce relational features between semantic units (a DOM node in our case).
A similar idea of encoding word-pairs as vectors also show promising results for reasoning about cross-sentence relations in natural language processing~\cite{Joshi2018pair2vecCW}.
These together help us eliminate the need for human effort in both designing features and metric algorithms.
The learned website-invariant neural features further make our model more transferable.
Recent advances in natural language processing also show an emerging interest in learning powerful deep neural features for richly formatted texts and semi-structured data~\cite{Qian2018GraphIEAG, kocayusufoglu2019riser}.
Our work agrees with the findings in Fonduer~\cite{Wu2018FonduerKB}: neural networks can effectively replace hand-crafted features for extracting data in richly formatted documents.
To the best of our knowledge, we are among the first approach using neural networks for learning to represent DOM nodes in web pages to solve structured prediction.


%% file: sec_8_supp.tex
\appendix
\section{Appendix} 
\label{sec:appendix}
Here we introduce more implementation details and some additional analysis of the proposed method in order to make our work more reproducible.

\subsection{Implementation Details.} 
\noindent
\textbf{Filtering Fixed DOM Nodes.} 
\label{ssec:vni}
As we know, detail pages in a website usually share a lot of common boilerplate (e.g. navigation texts, headers, footers, etc) where the textual nodes do not convey variable values of our interested fields~\cite{Gulhane2010ExploitingCR, Li2012ExploitingAR}. 
One can quickly eliminate them from consideration to reduce the number of nodes we are labeling in the model.
We propose a simple heuristic to utilize site-level information to filter nodes that are constant values in all pages of a website.
Specifically, we first collect all possible XPaths (i.e. node identifiers) of textual nodes in the website and then keep the number of different texts inside each node.

Then, we rank all nodes by these numbers, and take top-k of the XPaths with more than two different values as the \textbf{variable nodes} (we set k as 500 in the experiments).
This step can remove most of the page-invariant textual nodes which are irrelevant information for data extraction such as copyright information, website names, and etc.
The basic assumption behind this filtering is that we believe \textit{value nodes} are more likely to have more different texts inside different pages and at least show two different values.
In this way, we can significantly reduce our training and inference time without losing performance.

\subsection{Hyper-parameters}
We show our hyper-parameters and detailed model architectures. 
\subsubsection*{The Node Encoder Module}
\quad
First, the character-embedding dimension $\text{dim}_c=100$, the word embeddings  dimension  $\text{dim}_w=100$, the discrete feature embeddings is  $\text{dim}_d$ is 20 for the leaf type tag feature and 30 for the node value-checker features.
For the CNN network (i.e. the char-level word encoder), we use 50 filters and 3 as kernel size.
We set the LSTM of node text and preceding texts are both 100 dimensions.
Therefore, we learned 300d node vectors at the end of this stage.
We train the node encoder with epoch number as 10 and a batch size of 16. 
For our ablation studies with NL-based sequence tagging tasks, 
we use a 100d LSTM or a 100-fileter CNNs for the node tagger.
\subsubsection*{The Pair-level Relation Network Module}
\quad
The XPath and position embedding vectors are having 30 as dimension, and the XPath LSTM has a 100-dimension hidden size. The MLP for the pair classifier is also a 100d dense layer.
We train the relation network with epoch number as 10 and a batch size of 32. 
We empirically set the threshold of $N$ to be 1 and the $m$ to be $10$ for pair construction and label aggregation.

\subsubsection*{Shared training parameters.}\quad 
We use dropout layers at the end of both stages with a rate as 0.3 for avoid over-fitting. 
For the optimizer, we use Adam for both stages, with the learning\_rate=0.001,
    beta\_1=0.9,
    beta\_2=0.999, and 
    epsilon=1e-07.

\subsection{More Analysis of FreeDOM}


\smallskip
\subsubsection{Effect of site-level voting}
\begin{figure}[t]
	\centering
	\includegraphics[width=1\linewidth]{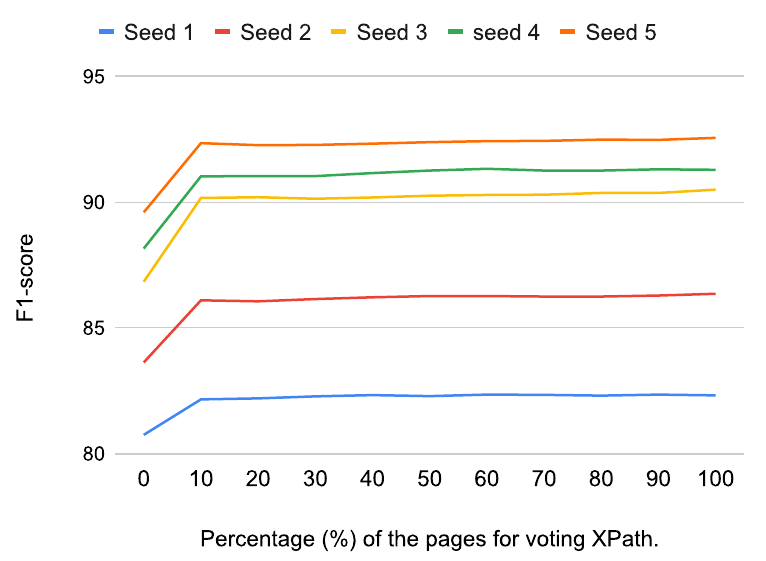}
	\caption{The curve of F1-score v.s. the percentage of the pages used for site-level voting, showing that even with streaming test data our methods still can achieve comparable performance under low-resource settings.}
	\label{fig:votingcurve}
\end{figure}

~\\
More specifically,
for each field $f_k$, we iterate all the perfections for $f_k$ in all the testing pages and then we rank the XPaths by their number of occurrence.
Then, in the second processing of each page, we rank all the XPaths in this page and order them with the previous computed occurrence scores, take the highest node as $f_k$ as well if it is labeled as other labels.
Fig.~\ref{fig:votingcurve} shows the curve when we use \textbf{different proportion} (0\%, 10\%, ..., 100\%) of the pages for voting XPaths.
The results with growing percentages used for majority voting can be a good estimation when the system is applied in online streaming scenarios where inputs are emerging unseen pages from different websites. 

\subsubsection{A closer look at a vertical}
We use an experiment in the \textit{Book} vertical using three seed websites ($k=3$) and compare the results from the first stage model and the second stage model.
Table~\ref{tab:case} clearly shows the improvement for each case when using our proposed pair-level networks for encoding relational information. 
The overall performance over these ten permutations of seed websites has a significant increment ({$88.90 \rightarrow 93.90$}). 
One of the largest improvements is from ``barnes.+bdep.+bsam.'' to ``abe.'' ($79.20 \rightarrow 98.60$).

\begin{figure*}[t]
	\centering
	\includegraphics[width=1\linewidth]{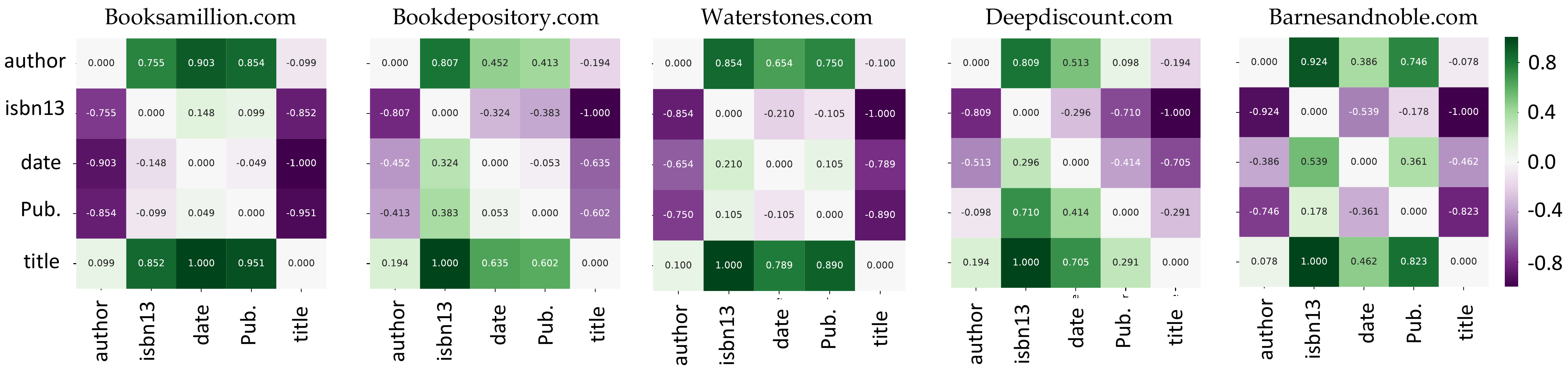}
	\caption{Patterns of distances between the value nodes of field-pairs in different websites of \textit{Book} vertical.
		The distance values are site-level normalized with MinMaxScaler, thus ranging form $-1$ to $+1$. Positive distances (in green) means the column-field is behind row-field. Similarly, the negative distances (in purple) mean that the row-fields are behind column-fields. The distance values in the figure are the number of variable nodes are there between two fields which are further normalized by the total number of nodes.
	}
	\label{fig:distviz}
\end{figure*}

We analyze the distance patterns of the \textit{abe}. pages and the three seed sites, and find that they share a similar inter-field patterns in (as our previous discussion about Figure~\ref{fig:distviz}).
This suggests that the second-stage module indeed captures the relational patterns.

\begin{table*}[h!]
	\scalebox{0.67
	}{
		\begin{tabular}{c|cccccccccc|c}
			\toprule
			\textbf{Seed \textbackslash~Target Site}     & \textit{abe.} & \textit{amazon} & \textit{barnes.} & \textit{bdep.} & \textit{bsam.} & \textit{borders} & \textit{buy} & \textit{chris.} & \textit{dee.} & \textit{water.} & \textbf{AVG}   \\ \midrule
			abe.+amazon+barnes.              & N/A     & N/A     & N/A     & 99.00 {\tiny{$\rightarrow$}}   99.00 & 99.67 {\tiny{$\rightarrow$}}   100.00  & 100.00 {\tiny{$\rightarrow$}}   100.00 & 66.67 {\tiny{$\rightarrow$}}   91.33 & 96.67 {\tiny{$\rightarrow$}}   96.67 & 100.00 {\tiny{$\rightarrow$}}   99.67  & 98.67 {\tiny{$\rightarrow$}}   98.67 & 94.38 {\tiny{$\rightarrow$}}   97.90 \\
			amazon+barnes.+bdep.        & 99.67 {\tiny{$\rightarrow$}}   99.67 & N/A     & N/A     & N/A     & 70.67 {\tiny{$\rightarrow$}}   78.67   & 99.67 {\tiny{$\rightarrow$}}   99.67   & 69.33 {\tiny{$\rightarrow$}}   91.00 & 97.33 {\tiny{$\rightarrow$}}   97.33 & 100.00 {\tiny{$\rightarrow$}}   100.00 & 99.00 {\tiny{$\rightarrow$}}   98.67 & 90.81 {\tiny{$\rightarrow$}}   95.00 \\
			barnes.+bdep.+bsam. & 78.80 {\tiny{$\rightarrow$}}   98.40 & 79.50 {\tiny{$\rightarrow$}}   84.50 & N/A     & N/A     & N/A       & 99.80 {\tiny{$\rightarrow$}}   99.80   & 82.20 {\tiny{$\rightarrow$}}   93.00 & 74.80 {\tiny{$\rightarrow$}}   94.00 & 98.00 {\tiny{$\rightarrow$}}   98.20   & 99.40 {\tiny{$\rightarrow$}}   98.60 & 87.50 {\tiny{$\rightarrow$}}   95.21 \\
			bdep.+bsam.+borders        & \underline{79.20 {\tiny{$\rightarrow$}}   98.60} & 78.25 {\tiny{$\rightarrow$}}   80.25 & 99.60 {\tiny{$\rightarrow$}}   99.40 & N/A     & N/A       & N/A       & 98.00 {\tiny{$\rightarrow$}}   95.00 & 75.80 {\tiny{$\rightarrow$}}   94.80 & 99.60 {\tiny{$\rightarrow$}}   99.40   & 99.60 {\tiny{$\rightarrow$}}   99.00 & 90.01 {\tiny{$\rightarrow$}}   95.21 \\
			bsam.+borders+buy                   & 78.60 {\tiny{$\rightarrow$}}   98.00 & 58.00 {\tiny{$\rightarrow$}}   63.75 & 99.60 {\tiny{$\rightarrow$}}   99.40 & 79.60 {\tiny{$\rightarrow$}}   82.00 & N/A       & N/A       & N/A     & 64.00 {\tiny{$\rightarrow$}}   93.00 & 92.00 {\tiny{$\rightarrow$}}   98.60   & 99.00 {\tiny{$\rightarrow$}}   98.60 & 81.54 {\tiny{$\rightarrow$}}   90.48 \\
			borders+buy+chris.                   & 91.40 {\tiny{$\rightarrow$}}   92.00 & 74.50 {\tiny{$\rightarrow$}}   95.00 & 78.80 {\tiny{$\rightarrow$}}   94.00 & 98.60 {\tiny{$\rightarrow$}}   97.40 & 97.40 {\tiny{$\rightarrow$}}   97.00   & N/A       & N/A     & N/A     & 99.00 {\tiny{$\rightarrow$}}   98.60   & 99.00 {\tiny{$\rightarrow$}}   98.80 & 91.24 {\tiny{$\rightarrow$}}   96.11 \\
			buy+chris.+dee.              & 90.20 {\tiny{$\rightarrow$}}   90.20 & 74.50 {\tiny{$\rightarrow$}}   74.50 & 79.60 {\tiny{$\rightarrow$}}   82.60 & 99.00 {\tiny{$\rightarrow$}}   98.20 & 59.20 {\tiny{$\rightarrow$}}   81.40   & 79.40 {\tiny{$\rightarrow$}}   99.00   & N/A     & N/A     & N/A       & 99.60 {\tiny{$\rightarrow$}}   99.00 & 83.07 {\tiny{$\rightarrow$}}   89.27 \\
			chris.+dee.+water.      & 97.40 {\tiny{$\rightarrow$}}   97.40 & 69.25 {\tiny{$\rightarrow$}}   75.75 & 74.20 {\tiny{$\rightarrow$}}   96.40 & 96.60 {\tiny{$\rightarrow$}}   94.00 & 50.20 {\tiny{$\rightarrow$}}   80.40   & 99.40 {\tiny{$\rightarrow$}}   98.80   & 77.60 {\tiny{$\rightarrow$}}   80.40 & N/A     & N/A       & N/A     & 80.66 {\tiny{$\rightarrow$}}   89.02 \\
			dee.+water.+abe.           & N/A     & 71.75 {\tiny{$\rightarrow$}}   77.50 & 79.60 {\tiny{$\rightarrow$}}   99.40 & 98.20 {\tiny{$\rightarrow$}}   87.80 & 99.00 {\tiny{$\rightarrow$}}   99.00   & 99.80 {\tiny{$\rightarrow$}}   99.80   & 95.80 {\tiny{$\rightarrow$}}   90.40 & 97.40 {\tiny{$\rightarrow$}}   97.60 & N/A       & N/A     & 91.65 {\tiny{$\rightarrow$}}   93.07 \\
			water.+abe.+amazon                 & N/A     & N/A     & 98.67 {\tiny{$\rightarrow$}}   99.00 & 94.33 {\tiny{$\rightarrow$}}   95.00 & 100.00 {\tiny{$\rightarrow$}}   100.00 & 99.67 {\tiny{$\rightarrow$}}   99.67   & 97.00 {\tiny{$\rightarrow$}}   93.67 & 97.67 {\tiny{$\rightarrow$}}   97.67 & 99.33 {\tiny{$\rightarrow$}}   99.33   & N/A     & 98.10 {\tiny{$\rightarrow$}}   97.76 \\ \midrule
			\textbf{AVG}                                         & 87.90 {\tiny{$\rightarrow$}}   96.32 & 72.25 {\tiny{$\rightarrow$}}   78.75 & 87.15 {\tiny{$\rightarrow$}}   95.74 & 95.05 {\tiny{$\rightarrow$}}   93.34 & 82.31 {\tiny{$\rightarrow$}}   90.92   & 96.82 {\tiny{$\rightarrow$}}   99.53   & 83.80 {\tiny{$\rightarrow$}}   90.69 & 86.24 {\tiny{$\rightarrow$}}   95.87 & 98.28 {\tiny{$\rightarrow$}}   99.11   & 99.18 {\tiny{$\rightarrow$}}   98.76 & \textbf{88.90 {\tiny{$\rightarrow$}}  93.90}\\ \bottomrule ~\\
		\end{tabular}
	}
	
	\caption{A detailed case study in the Book vertical for comparing the first-stage and second-stage model prediction results when $k=3$. Each entry in the table shows the improvement from first-stage results to ({{$\rightarrow$}}) second-stage ones.}
	\label{tab:case}
\end{table*}

%% file: main.bbl
\begin{thebibliography}{47}
\providecommand{\natexlab}[1]{#1}
\providecommand{\url}[1]{\texttt{#1}}
\expandafter\ifx\csname urlstyle\endcsname\relax
  \providecommand{\doi}[1]{doi: #1}\else
  \providecommand{\doi}{doi: \begingroup \urlstyle{rm}\Url}\fi

\bibitem[Azir and Ahmad(2017)]{Azir2017WrapperAF}
Mohd Amir Bin~Mohd Azir and Kamsuriah Ahmad.
\newblock Wrapper approaches for web data extraction : A review.
\newblock pages 1--6, 2017.

\bibitem[Carlson and Schafer(2008)]{carlson2008bootstrapping}
Andrew Carlson and Charles Schafer.
\newblock Bootstrapping information extraction from semi-structured web pages.
\newblock In \emph{Joint European Conference on Machine Learning and Knowledge
  Discovery in Databases}, pages 195--210. Springer, 2008.

\bibitem[Chang et~al.(2006)Chang, Kayed, Girgis, and Shaalan]{Chang2006ASO}
Chia-Hui Chang, Mohammed~O. Kayed, Moheb~R. Girgis, and Khaled~F. Shaalan.
\newblock A survey of web information extraction systems.
\newblock \emph{IEEE Transactions on Knowledge and Data Engineering},
  18:\penalty0 1411--1428, 2006.

\bibitem[Cohen et~al.(2015)Cohen, Ding, and
  Bagherjeiran]{Cohen2015SemiSupervisedWW}
Joseph~Paul Cohen, Wei Ding, and Abraham Bagherjeiran.
\newblock Semi-supervised web wrapper repair via recursive tree matching.
\newblock \emph{ArXiv}, abs/1505.01303, 2015.

\bibitem[Crescenzi et~al.(2001)Crescenzi, Mecca, and
  Merialdo]{Crescenzi2001RoadRunnerTA}
Valter Crescenzi, Giansalvatore Mecca, and Paolo Merialdo.
\newblock Roadrunner: Towards automatic data extraction from large web sites.
\newblock In \emph{VLDB}, 2001.

\bibitem[Cui et~al.(2017)Cui, Xiao, Wang, Song, won Hwang, and
  Wang]{Cui2017KBQALQ}
Wanyun Cui, Yanghua Xiao, Haixun Wang, Yangqiu Song, Seung won Hwang, and
  Wei~Yang Wang.
\newblock Kbqa: Learning question answering over qa corpora and knowledge
  bases.
\newblock \emph{PVLDB}, 10:\penalty0 565--576, 2017.

\bibitem[Devlin et~al.(2019)Devlin, Chang, Lee, and
  Toutanova]{Devlin2019BERTPO}
Jacob Devlin, Ming-Wei Chang, Kenton Lee, and Kristina Toutanova.
\newblock Bert: Pre-training of deep bidirectional transformers for language
  understanding.
\newblock In \emph{NAACL-HLT}, 2019.

\bibitem[Dong et~al.(2014{\natexlab{a}})Dong, Gabrilovich, Heitz, Horn, Lao,
  Murphy, Strohmann, Sun, and Zhang]{Dong2014KnowledgeVA}
Xin Dong, Evgeniy Gabrilovich, Geremy Heitz, Wilko Horn, Ni~Lao, Kevin Murphy,
  Thomas Strohmann, Shaohua Sun, and Wei Zhang.
\newblock Knowledge vault: a web-scale approach to probabilistic knowledge
  fusion.
\newblock In \emph{KDD}, 2014{\natexlab{a}}.

\bibitem[Dong et~al.(2014{\natexlab{b}})Dong, Gabrilovich, Heitz, Horn, Murphy,
  Sun, and Zhang]{Dong2014FromDF}
Xin Dong, Evgeniy Gabrilovich, Geremy Heitz, Wilko Horn, Kevin Murphy, Shaohua
  Sun, and Wei Zhang.
\newblock From data fusion to knowledge fusion.
\newblock \emph{PVLDB}, 7:\penalty0 881--892, 2014{\natexlab{b}}.

\bibitem[Ferrara et~al.(2014)Ferrara, De~Meo, Fiumara, and
  Baumgartner]{ferrara2014web}
Emilio Ferrara, Pasquale De~Meo, Giacomo Fiumara, and Robert Baumgartner.
\newblock Web data extraction, applications and techniques: A survey.
\newblock \emph{Knowledge-based systems}, 70:\penalty0 301--323, 2014.

\bibitem[Gentile et~al.(2013)Gentile, Zhang, and Ciravegna]{Gentile2013WebSI}
Anna~Lisa Gentile, Ziqi Zhang, and Fabio Ciravegna.
\newblock Web scale information extraction with lodie.
\newblock In \emph{AAAI Fall Symposia}, 2013.

\bibitem[Gulhane et~al.(2010)Gulhane, Rastogi, Sengamedu, and
  Tengli]{Gulhane2010ExploitingCR}
Pankaj Gulhane, Rajeev Rastogi, Srinivasan~H. Sengamedu, and Ashwin Tengli.
\newblock Exploiting content redundancy for web information extraction.
\newblock In \emph{WWW}, 2010.

\bibitem[Gulhane et~al.(2011)Gulhane, Madaan, Mehta, Ramamirtham, Rastogi,
  Satpal, Sengamedu, Tengli, and Tiwari]{Gulhane2011WebscaleIE}
Pankaj Gulhane, Amit Madaan, Rupesh~R. Mehta, Jeyashankher Ramamirtham, Rajeev
  Rastogi, Sandeepkumar Satpal, Srinivasan~H. Sengamedu, Ashwin Tengli, and
  Charu Tiwari.
\newblock Web-scale information extraction with vertex.
\newblock \emph{2011 IEEE 27th International Conference on Data Engineering},
  pages 1209--1220, 2011.

\bibitem[Hao et~al.(2011)Hao, Cai, Pang, and Zhang]{Hao2011FromOT}
Qiang Hao, Rui Cai, Yanwei Pang, and Lei Zhang.
\newblock From one tree to a forest: a unified solution for structured web data
  extraction.
\newblock In \emph{SIGIR}, 2011.

\bibitem[Joshi et~al.(2019)Joshi, Choi, Levy, Weld, and
  Zettlemoyer]{Joshi2018pair2vecCW}
Mandar Joshi, Eunsol Choi, Omer Levy, Daniel Weld, and Luke Zettlemoyer.
\newblock pair2vec: Compositional word-pair embeddings for cross-sentence
  inference.
\newblock In \emph{Proceedings of NAACL-HLT}, 2019.

\bibitem[Kocayusufoglu et~al.(2019)Kocayusufoglu, Sheng, Vo, Wendt, Zhao, Tata,
  and Najork]{kocayusufoglu2019riser}
Furkan Kocayusufoglu, Ying Sheng, Nguyen Vo, James Wendt, Qi~Zhao, Sandeep
  Tata, and Marc Najork.
\newblock Riser: Learning better representations for richly structured emails.
\newblock In \emph{The World Wide Web Conference}, pages 886--895. ACM, 2019.

\bibitem[Kushmerick et~al.(1997)Kushmerick, Weld, and
  Doorenbos]{Kushmerick1997WrapperIF}
Nicholas Kushmerick, Daniel~S. Weld, and Robert~B. Doorenbos.
\newblock Wrapper induction for information extraction.
\newblock In \emph{IJCAI}, 1997.

\bibitem[Lafferty et~al.(2001)Lafferty, McCallum, and
  Pereira]{Lafferty2001ConditionalRF}
John~D. Lafferty, Andrew McCallum, and Fernando Pereira.
\newblock Conditional random fields: Probabilistic models for segmenting and
  labeling sequence data.
\newblock In \emph{ICML}, 2001.

\bibitem[Lample et~al.(2016)Lample, Ballesteros, Subramanian, Kawakami, and
  Dyer]{Lample2016NeuralAF}
Guillaume Lample, Miguel Ballesteros, Sandeep Subramanian, Kazuya Kawakami, and
  Chris Dyer.
\newblock Neural architectures for named entity recognition.
\newblock In \emph{HLT-NAACL}, 2016.

\bibitem[Li et~al.(2012)Li, Zhu, Yin, Wang, and Wang]{Li2012ExploitingAR}
Xiang Li, Yanxu Zhu, Gang Yin, Tao Wang, and Huaimin Wang.
\newblock Exploiting attribute redundancy in extracting open source forge
  websites.
\newblock \emph{2012 International Conference on Cyber-Enabled Distributed
  Computing and Knowledge Discovery}, pages 13--20, 2012.

\bibitem[Lin et~al.(2017)Lin, Xu, Luo, and Zhu]{Lin2017MultichannelBM}
Bill~Y. Lin, Frank~F. Xu, Zhiyi Luo, and Kenny~Q. Zhu.
\newblock Multi-channel bilstm-crf model for emerging named entity recognition
  in social media.
\newblock In \emph{NUT@EMNLP}, 2017.

\bibitem[Lin and Lu(2018)]{Lin2018NeuralAL}
Bill~Yuchen Lin and Wei Lu.
\newblock Neural adaptation layers for cross-domain named entity recognition.
\newblock In \emph{EMNLP}, 2018.

\bibitem[Lin et~al.(2019{\natexlab{a}})Lin, Chen, Chen, and
  Ren]{lin-etal-2019-kagnet}
Bill~Yuchen Lin, Xinyue Chen, Jamin Chen, and Xiang Ren.
\newblock {K}ag{N}et: Knowledge-aware graph networks for commonsense reasoning.
\newblock In \emph{Proceedings of the 2019 Conference on Empirical Methods in
  Natural Language Processing and the 9th International Joint Conference on
  Natural Language Processing}, 2019{\natexlab{a}}.

\bibitem[Lin et~al.(2019{\natexlab{b}})Lin, Lee, Xu, Lan, and
  Ren]{lin-etal-2019-alpacatag}
Bill~Yuchen Lin, Dong-Ho Lee, Frank~F. Xu, Ouyu Lan, and Xiang Ren.
\newblock {A}lpaca{T}ag: An active learning-based crowd annotation framework
  for sequence tagging.
\newblock In \emph{Proceedings of the 57th Annual Meeting of the Association
  for Computational Linguistics: System Demonstrations}, 2019{\natexlab{b}}.

\bibitem[Lockard et~al.(2018)Lockard, Dong, Shiralkar, and Einolghozati]{ceres}
Colin Lockard, Xin~Luna Dong, Prashant Shiralkar, and Arash Einolghozati.
\newblock {CERES:} distantly supervised relation extraction from the
  semi-structured web.
\newblock \emph{{PVLDB}}, 11\penalty0 (10):\penalty0 1084--1096, 2018.
\newblock \doi{10.14778/3231751.3231758}.

\bibitem[Lockard et~al.(2020)Lockard, Shiralkar, Dong, and
  Hajishirzi]{Lockard2020ZeroShotCeresZR}
Colin Lockard, Prashant Shiralkar, Xin Dong, and Hannaneh Hajishirzi.
\newblock Zeroshotceres: Zero-shot relation extraction from semi-structured
  webpages.
\newblock In \emph{Proceedings of ACL}, 2020.

\bibitem[Luo et~al.(2018)Luo, Huang, Xu, Lin, Shi, and Zhu]{Luo2018ExtRAEP}
Zhiyi Luo, Shanshan Huang, Frank~F. Xu, Bill~Yuchen Lin, Hanyuan Shi, and
  Kenny~Q. Zhu.
\newblock Extra: Extracting prominent review aspects from customer feedback.
\newblock In \emph{EMNLP}, 2018.

\bibitem[Ma et~al.(2019)Ma, Zhang, Cao, Jin, Wang, Liu, Ma, and
  Ren]{Ma2019JointlyLE}
Weizhi Ma, Min Zhang, Yue Cao, Woojeong Jin, Chenyang Wang, Yiqun Liu, Shaoping
  Ma, and Xiang Ren.
\newblock Jointly learning explainable rules for recommendation with knowledge
  graph.
\newblock In \emph{WWW}, 2019.

\bibitem[Ma and Hovy(2016)]{ma2016end}
Xuezhe Ma and Eduard Hovy.
\newblock End-to-end sequence labeling via bi-directional lstm-cnns-crf.
\newblock \emph{arXiv preprint arXiv:1603.01354}, 2016.

\bibitem[Melnik et~al.(2011)Melnik, Gubarev, Long, Romer, Shivakumar, Tolton,
  and Vassilakis]{Melnik2011DremelIA}
Sergey Melnik, Andrey Gubarev, Jing~Jing Long, Geoffrey Romer, Shiva
  Shivakumar, Matt Tolton, and Theo Vassilakis.
\newblock Dremel: interactive analysis of web-scale datasets.
\newblock \emph{Commun. ACM}, 54:\penalty0 114--123, 2011.

\bibitem[Omari et~al.(2016)Omari, Kimelfeld, Yahav, and
  Shoham]{omari2016lossless}
Adi Omari, Benny Kimelfeld, Eran Yahav, and Sharon Shoham.
\newblock Lossless separation of web pages into layout code and data.
\newblock In \emph{Proceedings of the 22nd ACM SIGKDD international conference
  on knowledge discovery and data mining}, pages 1805--1814, 2016.

\bibitem[Pennington et~al.(2014)Pennington, Socher, and
  Manning]{pennington2014glove}
Jeffrey Pennington, Richard Socher, and Christopher~D. Manning.
\newblock Glove: Global vectors for word representation.
\newblock In \emph{Empirical Methods in Natural Language Processing (EMNLP)},
  pages 1532--1543, 2014.

\bibitem[Prasha~Shresha and Volkova(2019)]{mlg2019_45}
Dustin~Arendt Prasha~Shresha, Suraj~Maharjan and Svitlana Volkova.
\newblock Forecasting social interactions from dynamic graphs: A case study of
  twitter, github, and youtube.
\newblock In \emph{Proceedings of the 15th International Workshop on Mining and
  Learning with Graphs (MLG)}, 2019.

\bibitem[Qian et~al.(2018)Qian, Santus, Jin, Guo, and
  Barzilay]{Qian2018GraphIEAG}
Yujie Qian, Enrico Santus, Zhijing Jin, Jiang Guo, and Regina Barzilay.
\newblock Graphie: A graph-based framework for information extraction.
\newblock In \emph{NAACL-HLT}, 2018.

\bibitem[Ren et~al.(2016)Ren, He, Qu, Huang, Ji, and Han]{Ren2016AFETAF}
Xiang Ren, Wenqi He, Meng Qu, Lifu Huang, Heng Ji, and Jiawei Han.
\newblock Afet: Automatic fine-grained entity typing by hierarchical
  partial-label embedding.
\newblock In \emph{EMNLP}, 2016.

\bibitem[Santoro et~al.(2017)Santoro, Raposo, Barrett, Malinowski, Pascanu,
  Battaglia, and Lillicrap]{Santoro2017ASN}
Adam Santoro, David Raposo, David G.~T. Barrett, Mateusz Malinowski, Razvan
  Pascanu, Peter~W. Battaglia, and Timothy~P. Lillicrap.
\newblock A simple neural network module for relational reasoning.
\newblock In \emph{NIPS}, 2017.

\bibitem[Schulz et~al.(2016)Schulz, L{\"a}ssig, and
  Gaedke]{schulz2016practical}
Andreas Schulz, J{\"o}rg L{\"a}ssig, and Martin Gaedke.
\newblock Practical web data extraction: are we there yet?-a short survey.
\newblock In \emph{2016 IEEE/WIC/ACM International Conference on Web
  Intelligence (WI)}, pages 562--567. IEEE, 2016.

\bibitem[Sleiman and Corchuelo(2013)]{Sleiman2013ASO}
Hassan~A. Sleiman and Rafael Corchuelo.
\newblock A survey on region extractors from web documents.
\newblock \emph{IEEE Transactions on Knowledge and Data Engineering},
  25:\penalty0 1960--1981, 2013.

\bibitem[Talmor and Berant(2018)]{talmor-berant-2018-web}
Alon Talmor and Jonathan Berant.
\newblock The web as a knowledge-base for answering complex questions.
\newblock In \emph{Proceedings of the 2018 Conference of the North {A}merican
  Chapter of the Association for Computational Linguistics: Human Language
  Technologies, Volume 1 (Long Papers)}, pages 641--651, New Orleans,
  Louisiana, June 2018. Association for Computational Linguistics.
\newblock \doi{10.18653/v1/N18-1059}.

\bibitem[Trivedi et~al.(2017)Trivedi, Dai, Wang, and
  Song]{Trivedi2017KnowEvolveDT}
Rakshit Trivedi, Hanjun Dai, Yichen Wang, and Le~Song.
\newblock Know-evolve: Deep temporal reasoning for dynamic knowledge graphs.
\newblock In \emph{ICML}, 2017.

\bibitem[Wang et~al.(2019)Wang, Zhang, Zhao, Li, Xie, and
  Guo]{Wang2019MultiTaskFL}
Hongwei Wang, Fuzheng Zhang, Miao Zhao, Wenjie Li, Xing Xie, and Minyi Guo.
\newblock Multi-task feature learning for knowledge graph enhanced
  recommendation.
\newblock In \emph{WWW}, 2019.

\bibitem[Wu et~al.(2018)Wu, Hsiao, Cheng, Hancock, Rekatsinas, Levis, and
  R{\'e}]{Wu2018FonduerKB}
Sen Wu, Luke Hsiao, Xiao Cheng, Braden Hancock, Theodoros Rekatsinas, Philip
  Levis, and Christopher R{\'e}.
\newblock Fonduer: Knowledge base construction from richly formatted data.
\newblock \emph{Proceedings of the 2018 International Conference on Management
  of Data}, 2018.

\bibitem[Yang et~al.(2016)Yang, Yang, Dyer, He, Smola, and
  Hovy]{Yang2016HierarchicalAN}
Zichao Yang, Diyi Yang, Chris Dyer, Xiaodong He, Alexander~J. Smola, and
  Eduard~H. Hovy.
\newblock Hierarchical attention networks for document classification.
\newblock In \emph{HLT-NAACL}, 2016.

\bibitem[Zeng et~al.(2014)Zeng, Liu, Lai, Zhou, and Zhao]{Zeng2014RelationCV}
Daojian Zeng, Kang Liu, Siwei Lai, Guangyou Zhou, and Jun Zhao.
\newblock Relation classification via convolutional deep neural network.
\newblock In \emph{COLING}, 2014.

\bibitem[Zhai and Liu(2005)]{Zhai2005WebDE}
Yanhong Zhai and Bing Liu.
\newblock Web data extraction based on partial tree alignment.
\newblock In \emph{WWW}, 2005.

\bibitem[Zhou et~al.(2020)Zhou, Lin, Lin, Wang, Du, Neves, and Ren]{NERO2020}
Wenxuan Zhou, Hongtao Lin, Bill~Yuchen Lin, Ziqi Wang, Junyi Du, Leonardo
  Neves, and Xiang Ren.
\newblock Nero: A neural rule grounding framework for label-efficient relation
  extraction.
\newblock In \emph{Proceedings of The Web Conference 2020}, WWW ’20, page
  2166–2176, New York, NY, USA, 2020. Association for Computing Machinery.
\newblock ISBN 9781450370233.
\newblock \doi{10.1145/3366423.3380282}.

\bibitem[Zhu et~al.(2005)Zhu, Nie, Wen, Zhang, and Ma]{Zhu20052DCR}
Jun Zhu, Zaiqing Nie, Ji-Rong Wen, Bo~Zhang, and Wei-Ying Ma.
\newblock 2d conditional random fields for web information extraction.
\newblock In \emph{ICML}, 2005.

\end{thebibliography}
